\documentclass[10pt,journal,compsoc]{IEEEtran}
\pdfoutput=1
\usepackage{amsmath,amsfonts}
\usepackage{algorithmic}
\usepackage{array}
\usepackage[caption=false,font=normalsize,labelfont=sf,textfont=sf]{subfig}
\usepackage{textcomp}
\usepackage{stfloats}
\usepackage{url}
\usepackage{verbatim}
\usepackage{graphicx}
\usepackage{float}
\usepackage{subfig}
\usepackage{amsmath,amssymb,amsfonts}
\usepackage{color}
\usepackage{blindtext}
\usepackage{amsthm}
\usepackage[normalem]{ulem}
\usepackage{url}
\usepackage{comment}
\usepackage{multirow}
\usepackage{xcolor}
\usepackage{diagbox}
\usepackage{indentfirst}
\usepackage{array}
\usepackage[ruled]{algorithm2e}
\usepackage{algorithmic}
\usepackage{hyperref}
\usepackage{threeparttable}
\usepackage{balance}

\newcounter{gaocomm}
\newcounter{Note}
\definecolor{blue-violet}{rgb}{0.00,0.75,0.90}
\definecolor{mygreen}{rgb}{0.0, 0.5, 0.0}
\definecolor{awesome}{rgb}{1.0, 0.13, 0.32}
\definecolor{bostonuniversityred}{rgb}{0.8, 0.0, 0.0}
\ifCLASSOPTIONcompsoc
  \usepackage[nocompress]{cite}
\else
  \usepackage{cite}
\fi

\begin{document}
\title{A Survey on Temporal Knowledge Graph Completion: Taxonomy, Progress, and Prospects}

\author{Jiapu~Wang,
        ~Boyue~Wang,
        ~Meikang~Qiu, \IEEEmembership{Senior Member,~IEEE},
        ~Shirui~Pan,
        \IEEEmembership{Senior Member,~IEEE},\\
        ~Bo~Xiong,
        ~Heng~Liu,
        ~Linhao~Luo,
        ~Tengfei Liu,
        ~Yongli~Hu,
        ~Baocai~Yin,
        ~Wen~Gao, \IEEEmembership{Fellow,~IEEE}
\IEEEcompsocitemizethanks{\IEEEcompsocthanksitem Jiapu Wang, Boyue Wang, Heng Liu, Tengfei Liu, Yongli Hu and Baocai Yin are with Beijing Municipal Key Lab of Multimedia and Intelligent Software Technology, Beijing  Artificial  Intelligence Institute, Faculty of Information Technology, Beijing University of Technology, Beijing 100124, China. 
Email: \{jpwang, liuheng, tfliu\}@emails.bjut.edu.cn, \{wby, huyongli, ybc\}@bjut.edu.cn.
\IEEEcompsocthanksitem Meikang Qiu is with the Beacom College of Computer and Cyber Sciences, Dakota State University, Madison, South Dakota, USA.  Email: qiumeikang@ieee.org
\IEEEcompsocthanksitem Shirui Pan is with the School of Information and Communication Technology and Institute for Integrated and Intelligent Systems (IIIS), Griffith University, Queensland, Australia. 
Email: s.pan@griffith.edu.au.
\IEEEcompsocthanksitem Bo Xiong is with the Department of Computer Science, University of Stuttgart, 70569 Stuttgart, Germany. E-mail: bo.xiong@ipvs.uni-stuttgart.de.
\IEEEcompsocthanksitem Linhao Luo is with the Department of Data Science and AI, Monash University, Melbourne, Australia. E-mail: linhao.luo@monash.edu.
%\IEEEcompsocthanksitem Junbin Gao is with the Discipline of Business Analytics, The University of Sydney Business School, The University of Sydney, NSW 2006, Australia. E-mail: junbin.gao@sydney.edu.au.
\IEEEcompsocthanksitem Wen Gao is with the Institute of Digital Media, Peking University, Beijing 100871, China and also with the School of Electronic and Computer Engineering, Peking University Shenzhen Graduate School, Shenzhen 518055, China. E-mail:  wgao@pku.edu.cn. %\protect\\
\IEEEcompsocthanksitem Jiapu Wang and Boyue Wang contributed equally to this work. 
\IEEEcompsocthanksitem Corresponding author: Shirui Pan.
}}

% \author{
% \IEEEcompsocitemizethanks{\IEEEcompsocthanksitem \protect\\
% E-mail: }}


\markboth{Journal of \LaTeX\ Class Files,~Vol.~??, No.~??, August~20YY}%
{Shell \MakeLowercase{\textit{et al.}}: Bare Demo of IEEEtran.cls for Computer Society Journals}



\IEEEtitleabstractindextext{%
\begin{abstract}
    Temporal characteristics are prominently evident in a substantial volume of knowledge, which underscores the pivotal role of \textit{Temporal Knowledge Graphs (TKGs)} in both academia and industry. %Specifically, scholars are increasingly recognizing the significant impact of TKGs on downstream applications, leading to a growing focus on their integrity.
%As temporal knowledge graphs (TKGs) play an increasingly essential role in both academia and industry, their integrity has drawn much attention. %temporal information
However, TKGs often suffer from incompleteness for three main reasons: the continuous emergence of new knowledge, the weakness of the algorithm for extracting structured information from unstructured data, and the lack of information in the source dataset. 
Thus, the task of \textit{Temporal Knowledge Graph Completion (TKGC)} has attracted increasing attention, aiming to predict missing items based on the available information. In this paper, we provide a comprehensive review of TKGC methods and their details. Specifically, this paper mainly consists of three components, namely, \textit{1) Background,} which covers the preliminaries of TKGC methods, loss functions required for training, as well as the dataset and evaluation protocol; \textit{2) Interpolation,} that estimates and predicts the missing elements or set of elements through the relevant available information. It further categorizes related TKGC methods based on how to process temporal information; \textit{3) Extrapolation,} which typically focuses on continuous TKGs and predicts future events, and then classifies all extrapolation methods based on the algorithms they
utilize%; \textit{4) Applications,} which aim to explore TKGC in downstream applications, such as recommendation systems and risk analysis systems 
. We further pinpoint the challenges and discuss future research directions of TKGC.
\end{abstract}

}

\maketitle

%Adding the table of content to see the structure of the paper
%\tableofcontents

\IEEEdisplaynontitleabstractindextext

% \fi
\IEEEpeerreviewmaketitle

\IEEEraisesectionheading{\section{Introduction}\label{Sec:1}}
\IEEEPARstart{K}{nowledge} Graph (KGs) are structured multi-relational knowledge bases that typically contain a set of facts. Each fact in a KG is stored in the form of triplet $(\mathbf s,\ \mathbf r,\ \mathbf o)$, where $\mathbf s$ and $\mathbf o$ represent the head and tail entities, respectively, and $\mathbf r$ denotes the relation connecting the head entity and the tail entity. 
For example, given one triplet \textit{(Barack\ Hussein\ Obama,\ President\_of,\ USA)}, ``\textit{Barack\ Hussein\ Obama}" and ``\textit{USA}" are the head entity $\mathbf s$ and the tail entity $\mathbf o$, respectively, while ``\textit{president\_of}" represents the relation $\mathbf r$. Currently,  large-scale KGs are widely exploited in artificial intelligence and data mining applications, including traffic-flow forecasting \cite{9983531}, information retrieval \cite{Dalton2014EntityQF}, and dialogue systems \cite{Ma2015KnowledgeGI}.

Typically, the facts in KGs are time-specific and are valid only within a particular period, rendering knowledge time-limited. 
For example, the triplet \textit{(Barack\ Hussein\ Obama,\ President\_of,\ USA)} is only valid during the period \textit{[2009, 2017]}. Consequently, KGs that contain temporal labels form \textit{Temporal Knowledge Graphs (TKG)} and have gained significant attention in recent years. The fundamental unit of a TKG is a quadruplet $(\mathbf s,\ \mathbf r,\ \mathbf o,\ \mathbf t)$ formed by introducing the temporal information alongside the triplet. For example, the quadruplet can be represented as \textit{(Barack\ Hussein\ Obama,\ President\_of,\ USA,\ [2009, 2017])}. As events continually evolve, TKGs can be regularly updated to capture the dynamic changes in the real world. %providing research with valuable temporal information. %The example of TKG is shown in Fig. \ref{fig:my_label}.

\begin{comment}
    \begin{figure}[t]
    \begin{center}
    \includegraphics[scale=0.5]{Figures/Figure-TKG.png}
    \end{center}
    \caption{An example of the Temporal Knowledge Graph (TKG) and Temporal Knowledge Graph Completion (TKGC). The red bolded ``\textit{?} " represents the missing knowledge \textit{(Donald\ Trump,\ succeeded,\ Barack\ Hussein\ Obama,\ 2017)}.}
    \label{fig:my_label}
\end{figure}
\end{comment}

\begin{figure*}[t]
\subfloat[Interpolation]{
\includegraphics[scale=0.42]{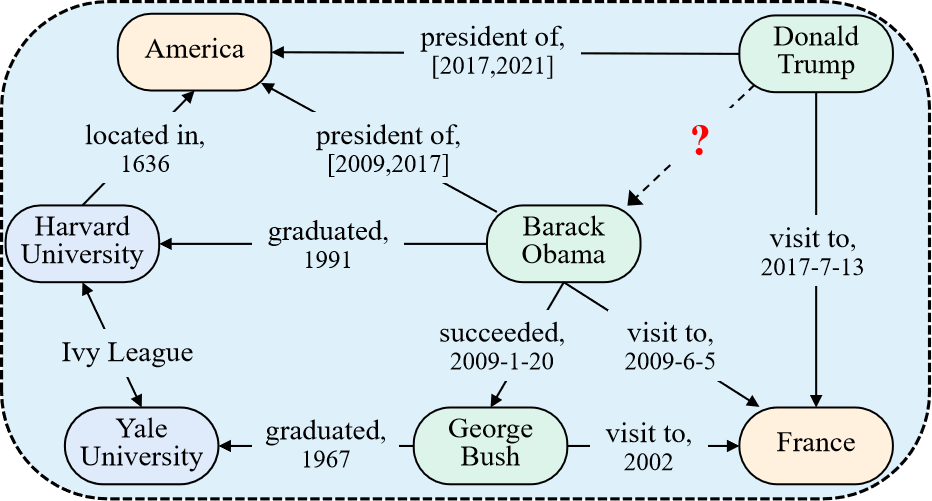}\label{fig:31}
}
\quad
\subfloat[Extrapolation]{
\includegraphics[scale=0.45]{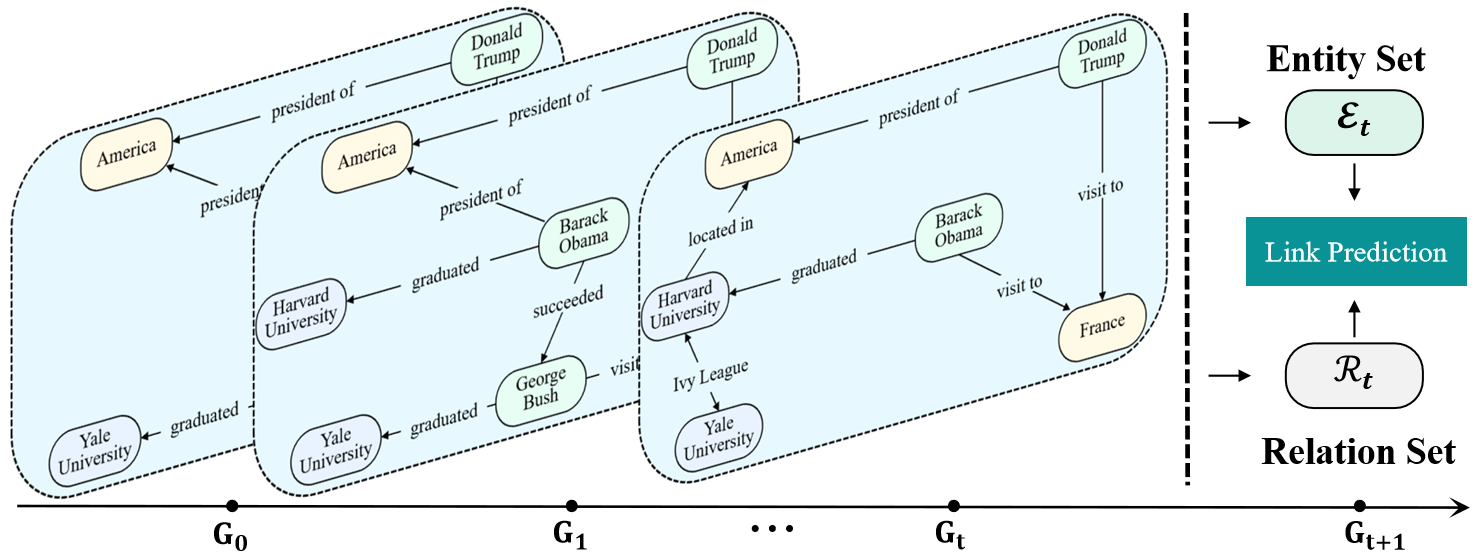}\label{fig:32}
}
\caption{Two categorizations of the \textit{Temporal Knowledge Graph Completion (TKGC)}, e.g., (a) Interpolation, in which the red bolded ``{\color{red}{\textbf{?}}} " represents the missing knowledge \textit{(Donald\ Trump,\ succeeded,\ Barack\ Hussein\ Obama,\ 2017)}; (b) Extrapolation, in which $\mathbf{G}_t$ denotes the static KG at time $t$.}
\label{fig:3}
\end{figure*} 

TKGs (\textit{e.g.,} ICEWS \cite{ICEWS2015} and GDELT \cite{leetaru2013gdelt}) can contain millions or even billions of quadruplets. However, despite their vastness, these TKGs remain incomplete for three primary reasons. Firstly, the source datasets often lack comprehensive information. Secondly, the algorithms used to extract entities and relations face challenges when dealing with diverse forms of unstructured data, rendering them less effective. Lastly, knowledge continuously evolves due to the ongoing events in nature. The incompleteness of the knowledge graph significantly hampers the effectiveness of knowledge-driven systems, thereby emphasizing the significance of \textit{Temporal Knowledge Graph Completion (TKGC)} as a crucial task.

The TKGC task aims to predict the missing items according to the available information, e.g., \textit{(Donald\ Trump,\ succeeded,\ Barack\ Hussein\ Obama,\ 2017)} can be inferred from existing quadruplets \textit{(Barack\ Hussein\ Obama,\ President\_of,\ USA,\ [2009, 2017])} and \textit{(Donald\ Trump,\ President\_of,\ USA,\ [2017, 2021])}.
Specifically, following the strategy whether forecasting future facts, we categorize existing TKGC literature into the \textit{interpolation methods} and the \textit{extrapolation methods}. More specifically, interpolation methods typically estimate unknown knowledge through the relevant known facts, and extrapolation methods aim to estimate unknown knowledge in the future. Fig \ref{fig:3} illustrates an example of these two different scenarios. %\SP{For each method, we need to use one sentence to explain/summarize it.} 

Each type of TKGC method faces specific important challenges that need to be addressed. 
When it comes to interpolation methods, two significant challenges arise: how to effectively incorporate temporal information into the evolution process of KGs and how to handle the timestamps to make full exploitation of its semantic information. Although some TKGC methods are designed to explore the semantics of temporal information, they often treat temporal information independently and fail to capture the hidden semantic information associated with the facts. Moreover, existing TKGC methods simply associate temporal information with the facts, which is challenging to reflect on the evolution process. On the other hand, extrapolation methods face the challenge of effectively mitigating the impact of anomalous historical information on TKG embedding. These methods explore the structural and temporal information among historical snapshots to further help predict future events. However, the presence of anomalous historical information severely hampers the accurate prediction of future events.

In response to these challenges, there have been increasing studies emerging recently. For instance, Ma~\emph{et al.} \cite{ma2021learning} employed bag-of-words and Bi-LSTM algorithms to fully explore the latent semantics of temporal information. Goel~\emph{et al.} \cite{goel2020diachronic} presented an approach called DE-SimplE, which incorporates a diachronic embedding function to integrate temporal information into facts. This approach effectively captures the evolution process of KGs, allowing for more accurate predictions. %DE-SimplE, and they proposed a diachronic embedding function to integrate the temporal information into facts, which can effectively reflect on the evolution process. Likewise, there are issues that have yet to be addressed and must be brought to the forefront. However, there is a lack of a comprehensive summary and comparison of existing methods.However, it is crucial to recognize that certain unresolved issues still demand attention and should be prioritized. 
However, it is crucial to recognize that there are still unresolved issues that require attention and should be prioritized. To facilitate further progress in the field, it becomes imperative to provide a comprehensive summary and comparison of existing TKGC methods. %Such an evaluation would serve as a valuable resource for researchers and practitioners, fostering a deeper understanding and facilitating advancements in the field of TKGC.
%To address the above challenges, a potential solution is to associate entities and relations to capture their hidden semantic information when encoding temporal information. In addition, we think applying many techniques to integrate temporal information into the entities and relations to enhance the evolution process of TKGs. Equally important is the further introduction of the adaptive strategy to weaken the effect of outliers in extrapolation methods. {\color{red}{This paragraph is the solution I proposed, but I don't know if it is needed. Writing it this way kind of looks like the challenge was solved.}} 

In this paper, we provide an overview of TKGC methods with a fine-grained categorization. We also summarize the benchmark datasets commonly used for the evaluation of TKGC methods and present the evaluation protocol. Furthermore, we analyze the challenges in the filed and discuss the future directions for this rapidly emerging topic. By conducting this comprehensive analysis, we aim to contribute to the advancement of TKGC research and provide insights for researchers and practitioners.
Our main contributions are summarized as follows:
\begin{enumerate}
    %\item \textbf{Comprehensive survey.} We systematically summarize all TKGC literature, especially the most recent ones. Equally important, we ensure that all literature types are appropriately proportioned. 
    \item \textbf{Comprehensive survey.} We conduct a systematic summary of all TKGC literature, with a particular focus on the most recent studies. %Equally important, we ensure that all literature types are appropriately proportioned. 
    Furthermore, we detail each TKGC method, make essential comparisons, and summarize the techniques and codes\footnote{https://github.com/jiapuwang/Awesome-TKGC} used.
    \item \textbf{Categorization and new taxonomies.} We provide a comprehensive summary and a fine-grained categorization of TKGCs. At the high level, we classify the current TKGC literature into two categories according to whether they forecast future events: the interpolation and the extrapolation methods. For interpolation methods, we divide them according to how to process temporal information. For the extrapolation methods, we classify them based on the algorithms they utilize.
    %\item \textbf{Summary of review.} We detail each classic method for each category of TKGCs, make essential comparisons, and summarize the techniques used.
    %\item \textbf{Comprehensive classification.} We classify the current TKGC literature into two categories according to whether they forecast future occurrences: the interpolation and the extrapolation methods. Afterward, we divide the two categories into multiple subcategories and further analyze this literature in depth.
    \item \textbf{Future directions.} We pinpoint future research directions of this fast-growing field, providing guidelines and suggestions on TKGC. 
\end{enumerate}

The rest of this paper is organized as follows. Section \ref{Sec:Background} briefly reviews the background of TKGC. Section \ref{sec:TKGC} details an overview and the categorization of TKGCs. In Section \ref{Sec:Interpolation}, we introduce the interpolation methods and classify in detail the interpolation methods. Likewise, Section \ref{Sec:Extrapolation} presents the extrapolation methods. Afterwards, we discuss the applications of TKG in recommendation and Q$\&$A systems in Section \ref{Sec:Applications}. Section \ref{Sec:Future} discusses the challenges and future research directions. Finally, we conclude this paper in Section \ref{Sec:Conclusion}.

\begin{table*}[t]
\centering
\caption{Statistic information of whole datasets.}
  \setlength{\tabcolsep}{1.1mm}{
  \begin{tabular}{c c c c c c c c c c c}
  \hline
  \rule{0pt}{9pt}
  {Datasets}&{\#Entities}&{\#Relations}&{\#Timestamps}&{\#Time Span}&{\#Training}&{\#Validation}&{\#Test}&{\#Granularity}&{\#Category}\\
  \hline
  \rule{0pt}{9pt}
  {ICEWS14 \cite{Learning2018}}&{6,869}&{230}&{365}&{01/01/2014 -- 12/31/2014}&{72,826}&{8,941}&{8,963}&{24 hours}&{Interpolation}\\
  \rule{0pt}{9pt}
  {ICEWS05-15 \cite{Learning2018}}&{10,094}&{251}&{4,017}&{01/01/2005 -- 12/31/2015}&{368,962}&{46,275}&{46,092}&{24 hours}&{Interpolation}\\
  %\rule{0pt}{8pt}
  %{ICEWS18 \cite{jin2019recurrent}}&{23,033}&{256}&{304}&{01/01/2018 - 12/31/2018}&{373,018}&{45,995}&{49,545}&{24 hours}&{Interpolation}\\
  \rule{0pt}{9pt}
  {GDELT \cite{leetaru2013gdelt}}&{500}&{20}&{366}&{04/01/2015 -- 03/31/2016}&{2,735,685}&{341,961}&{341,961}&{24 hours}&{Interpolation}\\
  \rule{0pt}{9pt}
  {YAGO11k \cite{dasgupta2018hyte}}&{10,623}&{10}&{70}&{-431 -- 2844}&{16,406}&{2,050}&{2,051}&{--}&{Interpolation}\\
  \rule{0pt}{9pt}
  {YAGO15k \cite{Learning2018}}&{15,403}&{34}&{198}&{1553 -- 2017}&{29,381}&{3,635}&{3,685}&{--}&{Interpolation}\\
  \rule{0pt}{9pt}
  {Wikidata12k \cite{dasgupta2018hyte}}&{12,554}&{24}&{81}&{1709 -- 2018}&{32,497}&{4,062}&{4,062}&{--}&{Interpolation}\\
  \hline
  \rule{0pt}{9pt}
  {ICEWS14 \cite{Learning2018}}&{6,869}&{230}&{365}&{01/01/2014 -- 12/31/2014}&{72,826}&{8,941}&{8,963}&{24 hours}&{Extrapolation}\\
  \rule{0pt}{8pt}
  {ICEWS18 \cite{jin2019recurrent}}&{23,033}&{256}&{304}&{01/01/2018 -- 10/31/2018}&{373,018}&{45,995}&{49,545}&{24 hours}&{Extrapolation}\\
  \rule{0pt}{9pt}
  {GDELT \cite{leetaru2013gdelt}}&{7,691}&{240}&{2,751}&{01/01/2018 -- 01/31/2018}&{1,734,399}&{238,765}&{305,241}&{15 mins}&{Extrapolation}\\
  \rule{0pt}{9pt}
  {WIKI \cite{leblay2018deriving}}&{12,554}&{24}&{232}&{1786 -- 2018}&{539,286}&{67,538}&{63,110}&{1 year}&{Extrapolation}\\
  \rule{0pt}{9pt}
  {YAGO \cite{mahdisoltani2014yago3}}&{10,623}&{10}&{189}&{1830 -- 2019}&{161,540}&{19,523}&{20,026}&{1 year}&{Extrapolation}\\
\hline
\end{tabular}
}\\[2mm]
\label{tab:label_data}
\end{table*}

\section{Background}\label{Sec:Background}
In this section, we describe the background of the \textit{Temporal Knowledge Graph Completion (TKGC)}, including \textit{Preliminaries}, \textit{Loss functions}, \textit{Benchmark datasets}, and \textit{Evaluation protocol}.

\subsection{Preliminaries}
\textit{Temporal Knowledge Graphs (TKGs)} are structured knowledge bases consisting of time-specific knowledge. Specifically, the TKG can be denoted as $\mathcal G = \{\mathcal{Q}\ |\ \mathcal E,\ \mathcal R,\ \mathcal{T}\}$, and these symbols $\mathcal E,\ \mathcal R,\ \mathcal T,\ \mathcal Q$ respectively represent the entity, relation, timestamp and quadruplet sets. More specifically, each knowledge in TKGs is stored in the form of quadruplet $(\mathbf{s},\ \mathbf{r},\ \mathbf{o},\ \mathbf{t})\in \mathcal Q$, where $\mathbf s,\ \mathbf o\in \mathcal E$ are the head and tail entities, $\mathbf r\in \mathcal R$ denotes the relation and $\mathbf t\in \mathcal T$ means the timestamp. Furthermore, $\mathbf s$ and $\mathbf o$ represent the nodes of the TKG, $\mathbf r$ denotes the edge from the head entity to the tail entity and $\mathbf t$ is the temporal label, which mainly contains two forms, such as the time point or the time interval.

\subsection{Loss Functions}

%Although existing TKGs have already contained millions or even billions of quadruplets, they are still far from complete due to the continuous emergence of new knowledge. 
Due to the incompleteness of TKGs, TKGC has become increasingly urgent. Specifically, in the TKGC task, the score function, denoted as $g(x)$, is introduced to evaluate or predict the credibility or probability of the quadruplet. Furthermore, the loss function $\mathcal{L}$ is designed to minimize the score function. Additionally, in order to enhance the efficiency of model training, negative sampling \cite{yang2020understanding} is incorporated into the TKGC task, whereby quadruplets are uniformly sampled from the entire set of possible quadruplets. In specific, we summarize two commonly-used loss functions and three temporal regularizations in the TKGC task. 

\textbf{Margin-based ranking loss} $\mathcal L_m$ \cite{bordes2013translating, jiang2016towards, leblay2018deriving} %is generally used in translation-based methods. The margin-based ranking loss 
introduces the margin to ensure that the score of the true quadruplet is lower than that of the corrupted quadruplet,
\begin{equation}
    \mathcal L_m =\sum_{(\mathbf s, \mathbf r,\mathbf o,\mathbf t)\in\mathcal{Q}}[\lambda+g(\mathbf s, \mathbf r,\mathbf o, \mathbf t)- \sum_{\mathbf o'\in \mathcal E}g(\mathbf s, \mathbf r, \mathbf o', \mathbf t)]_+,
\end{equation}
where $[x]_+=max(x,\ 0)$ and $\lambda>0$ is a margin hyperparameter. However, the margin-based ranking loss is sensitive to outliers. 

\textbf{Cross-entropy loss} $\mathcal L_c$ \cite{sadeghian2021chronor} calculates the difference between two probability distributions, accurately reflecting the prediction accuracy of the model while being less sensitive to outliers. In TKGC task, we follow the standard data augmentation protocol \cite{lacroix2018canonical} and add inverse relations to the datasets, i.e., creating one quadruplet $(\mathbf o,\ \mathbf r^{-1},\ \mathbf s,\ \mathbf t)$ for each quadruplet $(\mathbf s,\ \mathbf r,\ \mathbf o,\ \mathbf t)$. Afterwards, the cross-entropy loss can be defined as follows,
\begin{equation}
    \begin{split}
        \mathcal L_c 
        = & -\text{log}(\frac{\text{exp}(f(\mathbf s,\ \mathbf r,\ \mathbf o,\ \mathbf t))}{\sum_{\mathbf s'\in \mathcal E}\text{exp}(f(\mathbf s',\ \mathbf r,\ \mathbf o,\ \mathbf t))})\\
        & -\text{log}(\frac{\text{exp}(f(\mathbf o,\ \mathbf r^{-1},\ \mathbf s,\ \mathbf t))}{\sum_{\mathbf o'\in \mathcal E}\text{exp}(f(\mathbf o',\ \mathbf r^{-1},\ \mathbf s,\ \mathbf t))}).
    \end{split}
\end{equation}

%\subsection{Temporal Regularization} 
Finally, we summarize three classical temporal regularizations $\mathcal L_{\tau}$ widely used for ensuring facts behave smoothly over time in TKG. TComplEx \cite{lacroix2020tensor} first proposes the nuclear $3$-norm \cite{xu2021temporal} temporal regularization in the TKGC task, which expects corresponding elements of neighboring temporal embeddings to be close. The temporal regularization can be denoted as follows,
\begin{equation}
\label{LR}
    \mathcal L_\tau =\frac{1}{|\mathcal T|-1} \sum_{i=1}^{|\mathcal T|-1}\|\mathbf{t}_{i+1}-\mathbf{t}_i\|_3^3,
\end{equation}
where $|\mathcal T|$ is the number of timestamps and $\mathbf{t}_i$ denotes the $i$-th timestamp. Based on TComplEx, TeLM \cite{xu2021temporal} imposes a bias component into the temporal regularization, enhancing flexibility and expressiveness. Mathematically, this can be represented as follows,
\begin{equation}
    \mathcal L_\tau =\frac{1}{|\mathcal T|-1} \sum_{i=1}^{|\mathcal T|-1}\|\mathbf{t}_{i+1}-\mathbf{t}_i+\mathbf{t}_b\|_3^3,
\end{equation}
where $\mathbf{t}_b$ denotes the bias term and is randomly initialized.

However, the aforementioned temporal regularizations primarily emphasize the absolute distance between corresponding elements, thereby restricting the flexibility of neighboring temporal embeddings to a significant extent. QDN \cite{wang2023qdn} proposes a cosine similarity-based temporal regularization, which measures the global similarity between neighboring temporal embeddings,
\begin{equation}
\label{TR}
    \begin{split}
        \mathcal L_\tau = -\frac{1}{|\mathcal T|-1}\sum_{i=1}^{|\mathcal T|-1}(\mathbf{t}_{i+1}^T \cdot \mathbf{t}_i).
    \end{split}
\end{equation}

\begin{figure*}[t]
    \begin{center}
    \includegraphics[scale=0.55]{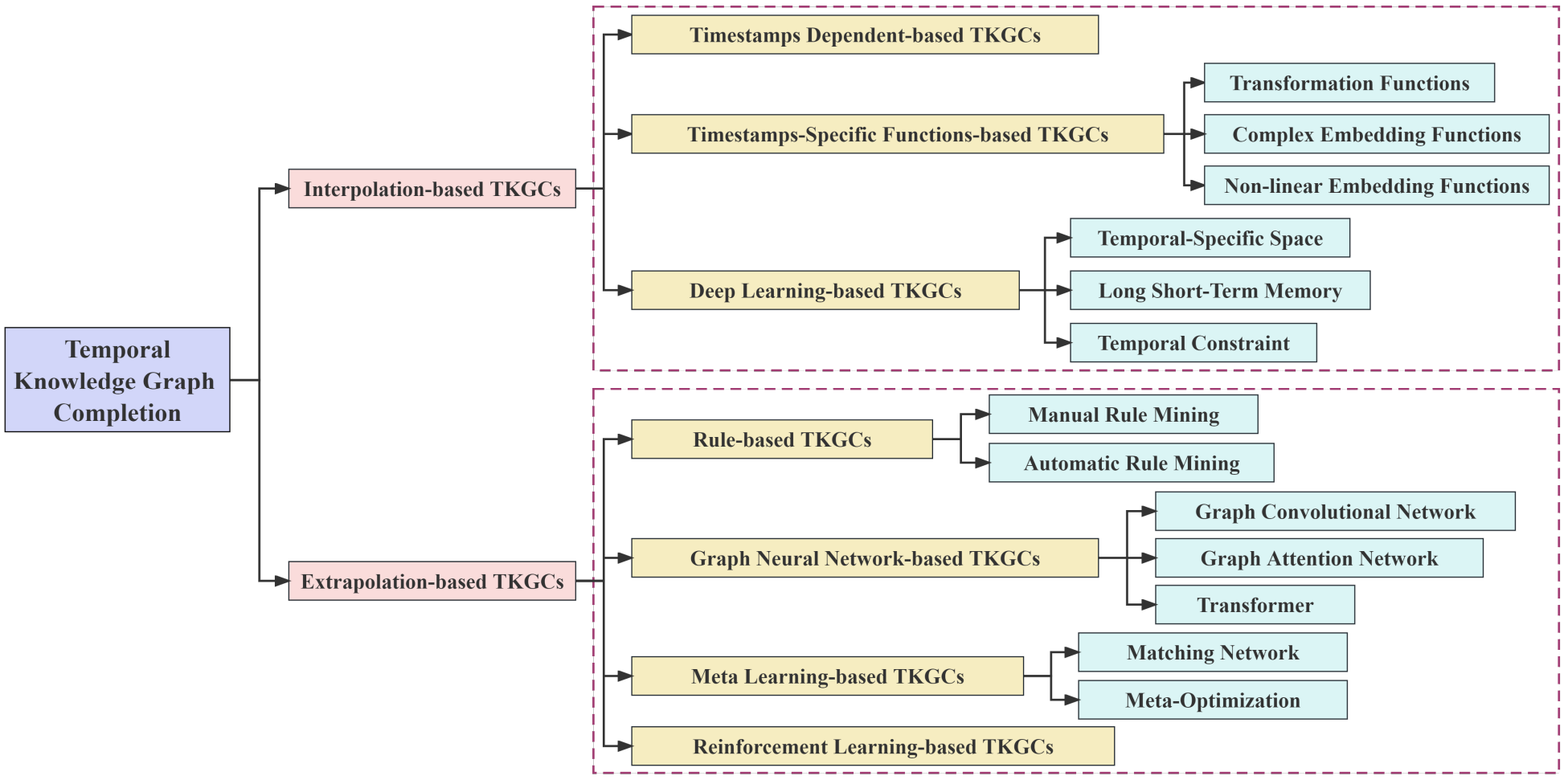}
    \end{center}
    \caption{Fine-grained categorization of \textit{Temporal Knowledge Graph Completion (TKGC)} methods.}
    \label{fig:my_label_1}
\end{figure*}

\subsection{Benchmark Datasets}
We list eleven common datasets widely used in the TKGC task and the major statistics of these datasets are summarized in TABLE \ref{tab:label_data} \cite{cai2022temporal}. \textbf{ICEWS14}, \textbf{ICEWS05-15} \cite{Learning2018} and \textbf{ICEWS18} \cite{jin2019recurrent} are three subsets of \textit{Integrated Crisis Early Warning System (ICEWS)} \cite{ICEWS2015}, which involves some political events with time points. \textbf{GDELT} \cite{leetaru2013gdelt} is a subset of the larger \textit{Global Database of Events, Language, and Tone (GDELT)} \cite{leetaru2013gdelt} that contains human social relationships, and timestamps are mainly in the form of time points. It should be emphasized that GDELT exhibits a complex geometry characterized by a small number of nodes and a substantial volume of training data.

\textbf{YAGO11k} \cite{dasgupta2018hyte}, \textbf{YAGO15k} \cite{Learning2018} and \textbf{YAGO} \cite{mahdisoltani2014yago3} are three subsets of YAGO3 \cite{mahdisoltani2014yago3}. \textbf{Wikidata12k} \cite{dasgupta2018hyte} and \textbf{WIKI} \cite{leblay2018deriving} are two subsets of WIKIDATA \cite{erxleben2014introducing}.
Different from ICEWS and GDELT, time annotations in YAGO11k and Wikidata12k are represented in various forms, i.e., time points like $2015-12-12$, time intervals like ``Since $2015$ ($[2015, \#\#]$)" and $[2015,\ 2016]$. Unlike ICEWS and GDELT, YAGO3 and WIKIDATA have a more sparse graph structure.

\subsection{Evaluation Protocol}
The evaluation protocol serves as a crucial metric for assessing the performance of TKGC methods. Typically, it involves replacing either the head or tail entity in each test quadruplet $(\mathbf s,\mathbf r,\mathbf o,\mathbf t)$ with all possible entities in the TKG, and subsequently ranking the scores produced by the scoring function. Some widely used evaluation metrics include Mean Rank (MR), Mean Reciprocal Rank (MRR) and Hit@$N$. 
\begin{itemize}
    \item \textbf{Mean Rank (MR)}: the average rank assigned to the true quadruplet overall test quadruplets;
    \item \textbf{Mean Reciprocal Rank (MRR)}: the average of the reciprocal rank assigned to the true quadruplet overall test quadruplets;
    \item \textbf{Hit@$N$}: the percentage of cases in which the true quadruplet appears in the top $N$ ranked quadruplets.  Here, we specifically report the $N = 1, 3, 10$ scores to verify the performance of TKGC methods.
\end{itemize}

Higher values of MRR and Hits@$N$, as well as lower MR, indicate better performance. In addition, the final experimental results consist of two flavors: \textit{Raw} and \textit{Filter}. Specifically, the \textit{Filter} metric is calculated by excluding all reconstituted quadruplets that existed in the training, validation, or test set from the rank, whereas the \textit{Raw} metric does not consider such exclusions. A more extensive description of these metrics can be found in \cite{trouillon2016complex, dettmers2018convolutional}.

\section{Temporal Knowledge Graph Completion}\label{sec:TKGC}

In this section, we %provide an overview of TKGCs and 
categorize them into two flavors based on whether they forecast future events, including \textit{Interpolation-based TKGCs} (Fig. \ref{fig:31}) and \textit{Extrapolation-based TKGCs} (Fig. \ref{fig:32}). The further fine-grained sub-categories are illustrated in Fig. \ref{fig:my_label_1}. 

%In this section, we provide an overview of TKGCs and categorize them into two flavors, including \textit{Interpolation} (Fig. \ref{fig:31}) and \textit{Extrapolation} (Fig. \ref{fig:32}). To better understand TKGs, we further detail the categorization of \textit{Interpolation} and \textit{Extrapolation}. The fine-grained categorization of TKGCs is illustrated in Fig. \ref{fig:my_label_1}. 

\subsection{Interpolation-based TKGCs}
Interpolation \cite{lunardi2009interpolation, gardner1993interpolation} is a statistical method %by which the relevant known values are used to estimate an unknown value or set of values. 
that use the relevant known values to estimate an unknown value or set. 
By identifying the consistent trend within a dataset, it becomes possible to reasonably estimate 
%\WangC{Maybe `update' or `optimize' is better than `estimate' here.} 
values that have not been explicitly calculated. 
%In the TKGC task, the interpolation methods generally aim to predict the missing item through the known knowledge in TKG. 
%\WangC{You have claimed this purpose of interpolation methods in section 3.} 

Interpolation-based TKGC methods generally complete the missing item by analyzing the known knowledge in TKGs. We categorize them based on how they process temporal information as follows,
\begin{enumerate}
    \item \emph{Timestamps dependent-based TKGC methods} do not impose operations on timestamps.
    \item \emph{Timestamps-specific functions-based TKGC methods} apply the timestamps-specific functions to obtain embeddings of timestamps or the evolution of entities and relations.
    \item \emph{Deep learning-based TKGC methods} utilize deep learning algorithms to encode temporal information and investigate the dynamic evolution of entities and relations.
\end{enumerate}

\subsection{Extrapolation-based TKGCs}

Extrapolation \cite{brezinski2013extrapolation, brezinski1980general} %, {\color{red}which is similar to interpolation to a certain extent},
focuses on forecasting the `future' unknown values beyond the data that is currently accessible.

%Extrapolation \cite{brezinski2013extrapolation, brezinski1980general} usually corresponds to interpolation, where extrapolation estimates unknown values beyond the known data. 
%In the TKGC task, the extrapolation methods typically focus on continuous TKG and predict future events.} 
%\WangC{This definition is the same as the Interpolation and cannot express the meaning of the future. How about ``Extrapolation \cite{brezinski2013extrapolation, brezinski1980general}, which is somewhat similar to interpolation, focus on forecasting the `future' unknown values beyond the data that is currently accessible.''} 

Extrapolation-based TKGC methods focus on continuous TKGs, enabling predictions of future events by learning embeddings of entities and relations from historical snapshots.
%learn embeddings of entities and relations from historical snapshots, thereby focusing on continuous TKGs to enable predictions of future events.
%forecast future events. 
We categorize them according to the algorithms they utilize as follows, %in predicting future events.
%{\color{blue}{classify them based on how they forecast future events.}} 
\begin{enumerate}
    \item \emph{Rule-based TKGC methods} apply the logical rules to reason %missing items. 
    future events.
    \item \emph{Graph neural network-based TKGC methods} generally utilize GNN and RNN to explore the structural and temporal information in TKG.
    \item \emph{Meta learning-based TKGC methods} design the meta-learner to instruct %the model on the process of learning.
    the learning process of the model. 
    \item \emph{Reinforcement learning-based TKGC methods} introduce the reinforcement learning strategy to ensure that the model achieves its training goals better.
\end{enumerate}

In the following sections (Section \ref{Sec:Interpolation} and Section \ref{Sec:Extrapolation}), we will introduce these TKGC categorizations in detail.

\section{Interpolation-based TKGCs} \label{Sec:Interpolation}
In this section, we provide an overview of the interpolation methods from three aspects: \textit{Timestamps dependent-based TKGC methods}, \textit{Timestamps-specific functions-based TKGC methods} and \textit{Deep learning-based TKGC methods}.

\subsection{Timestamps Dependent-based TKGCs}
\begin{figure}[t]
    \begin{center}
    \includegraphics[scale=0.56]{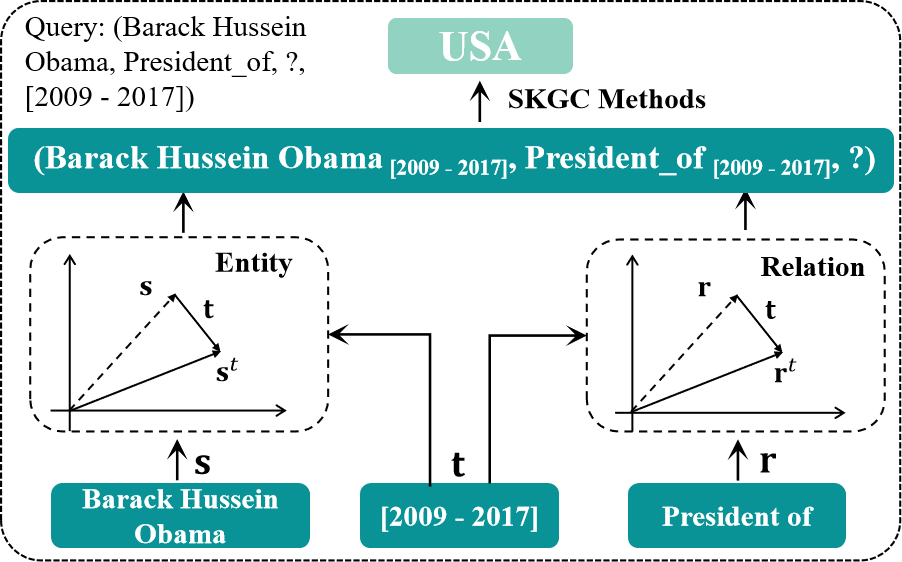}
    \end{center}
    \caption{The framework of the timestamps dependent-based TKGCs.}
    \label{fig:my_label0}
\end{figure}

Timestamps dependent-based TKGC methods typically %do not perform any operation on timestamps. 
do not perform operations on timestamps. 
Instead, they simply associate the timestamp with the corresponding entity or relation to accomplish the evolution of entities or relations. %Instead, they directly convert the quadruplet into a triplet to accomplish the evolution process of entities or relations by simply associating timestamps with entities and relations. 
As shown in Fig. \ref{fig:my_label0}, given a query \textit{(Barack Hussein Obama, President of, ?, [2009 - 2017])}, timestamps dependent-based TKGC methods generally associate the timestamp \textit{[2009 - 2017]} with the entity \textit{Barack Hussein Obama} and relation \textit{President of} to achieve the evolution process. 
Finally, they complete the missing item \textit{USA} using \textit{Static Knowledge Graph Completion (SKGC)} methods.

\textbf{TuckERTNT} \cite{shao2022tucker} is a classic timestamps dependent-based TKGC method that expands upon %builds upon the foundation of 
{TuckER} \cite{balavzevic2019tucker}. 
TuckER is modeled based on the triplet $(\mathbf{s},\ \mathbf{r},\ \mathbf{o})$ and incorporates a $3$rd-order tensor to facilitate the link prediction task. Here, TuckER explores the Tucker decomposition algorithm to overcome the over-parameterization problem. The score function of TuckER is defined as,
\begin{equation}
\begin{split}
    g(\mathbf{s},\ \mathbf{r},\ \mathbf{o})&= \mathcal{W} \times_1\mathbf{s} \times_2 \mathbf{r} \times_3 \mathbf{o},\\
    \mathcal{W}&= \mathcal{Z} \times_1\mathbf{A} \times_2 \mathbf{B} \times_3 \mathbf{C},
\end{split}
\end{equation}
where $\mathcal{W}$ denotes the $3$rd-order tensor, which can be decomposed as a core tensor $\mathcal{Z}$ and three factor matrices $\mathbf{A}$, $\mathbf{B}$ and $\mathbf{C}$; 
$\times_i$ represents the tensor multiplication between the core tensor and factor matrix in $i$-th dimension;
%$i$th-order tensor product; 
$g(\mathbf{s},\ \mathbf{r},\ \mathbf{o})$ is the score function. TuckERTNT is an extension of TuckER by introducing timestamps. Moreover, TuckERTNT takes into account the notion that certain facts may vary with respect to time, while others remain independent of time. It defines its score function as,
\begin{equation}
    g(\mathbf{s},\ \mathbf{r},\ \mathbf{o},\ \mathbf{t})= \mathcal{W} \times_1\mathbf{s} \times_2 (\mathbf{r} \circ \mathbf{t} +\mathbf{r}) \times_3 \mathbf{o},
\end{equation}
where $\circ$ denotes the Hadamard (or element-wise) product.

\textbf{TTransE} \cite{leblay2018deriving} extends the classic TransE by introducing the joint encoding of relations and timestamps within the same space.%, {\color{red}representing the translation from the head entity to the tail entity}. \WangC{We can delete this sentence.}  
Based on TTransE, \textbf{ST-TransE} \cite{9450214} further proposes a specific time embedding method to constrain the representation learning of entities and relations. 
Both TTransE and ST-TransE struggle to effectively handle facts that undergo temporal evolution.
\textbf{T-SimplE} \cite{9288194} introduces the $4$th-order tensor to capture the associativity among the elements of the quadruplet. Likewise,  \textbf{He} ~\emph{et al.} \cite{he2022improving} introduce \textit{Canonical Polyadic (CP)} decomposition into the TKGC task, and incorporate both temporal and non-temporal relations to model temporal facts (change over time) and non-temporal facts (not change over time). %\WangC{Sometime you use the algorithm name, while sometimes you use the person name. I tend to uniform them.} 
In particular, T-SimplE ignores the evolving strength of representations of pairwise relations in the same relational chain over time, as well as the revision of candidate prediction. Thus, \textbf{TKGFrame} \cite{zhang2020tkgframe} proposes a relation evolving enhanced model, which obtains more accurate TKG embeddings by learning a new temporal evolving matrix. 
Meanwhile, TKGFrame introduces a refinement model to revise the candidate predictions. To further capture the semantic property between relation and its involved entities at various time steps, \textbf{Li}~\emph{et al.} \cite{li2022leveraging} design several regularizations to constrain the expression of TKG. Recently, \textbf{TBDRI} \cite{yu2023tbdri} takes inverse relations as one of the most important types of relations. It independently learns inverse relations through block decomposition based on relational interaction. 

\subsection{Timestamps-specific Functions-based TKGCs}
Timestamps-specific functions-based TKGC methods exploit the specific functions to learn the embeddings of timestamps or the evolution of entities and relations,
%Timestamps-specific functions-based TKGC methods aim to learn the embeddings of timestamps through specific functions, 
such as diachronic embedding functions \cite{goel2020diachronic}, Gaussian functions \cite{xu2020temporal} and transformation functions \cite{zhao2023householder}. %\WangC{whether adds -s, i.e., transformation functions. Please update the function names.} 
%As shown in Fig. \ref{fig:my_label_ST}, the specific function-based TKGC methods embed the query \textit{(Barack Hussein Obama, President\_of, ?, [2009 - 2017])} into the hyperbolic space through the hyperbolic mapping function, and then explore the evolution process of entities and relations over time.

\subsubsection{Transformation Functions}
Transformation functions-based TKGC methods encode timestamps via the transformation functions. Some classic methods are proposed, such as \textbf{BoxTE} \cite{messner2022temporal}, \textbf{SPLIME} \cite{radstok2021leveraging} and \textbf{TARGCN} \cite{ding2021simple}.

\textbf{Jiang}~\emph{et al.} \cite{jiang2016towards} first propose a time-aware TKGC method to comprehensively capture the temporal nature of facts. Specifically, they  respectively utilize the time-aware embedding model and \textit{Integer Linear Programming (ILP)} to encode temporal order information and temporal consistency information. However, this method has room for improvement in terms of effectively leveraging temporal information. %and achieving a stronger model representation. 
\textbf{HTTR} \cite{zhao2023householder} introduces the Householder transformation \cite{de2002constrained} to explore the temporal evolution of relations. Specifically, HTTR defines the orthogonal matrix that represents the rotation from the head entity to the tail entity. This orthogonal matrix is obtained through the Householder transformation, and it directly links to the fused information of the relation and the temporal aspects.

\textbf{BoxTE} \cite{messner2022temporal} extends from %is a box embedding model for TKGC, building on 
the SKGC model BoxE \cite{abboud2020boxe}. 
BoxE encodes entities as points and represents relations as a set of boxes to enable the flexible representation of fundamental logical properties. 
Additionally, BoxTE embeds temporal information via the relation-specific transfer matrix to explore rich inference patterns. \textbf{Dai}~\emph{et al.} \cite{dai2022wasserstein} propose a model-agnostic method on the basis of BoxTE, in which they initially introduce the generative adversarial learning technique into the TKGC task. Specifically, the generator constructs high-quality plausible quadruplets and %to solve the problem of low-quality of negatively sampled samples by the traditional TKGC methods. Afterwards, 
the discriminator obtains the embeddings of entities and relations based on the generator. To overcome the problem of vanishing gradients on discrete data, Dai~\emph{et al.} simultaneously introduce the Wasserstein distance and Gumbel-Softmax relaxation.

\textbf{SPLIME} \cite{radstok2021leveraging} is essentially a transformation function, which applies \textit{splitting} and \textit{merging} operations to model TKGs through SKGC methods (e.g., McRL \cite{wang2023multi} and BoxE \cite{abboud2020boxe}). 
However, a substantial number of parameters introduced in SPLIME lead to large memory consumption and limit the running efficiency. 
\textbf{Ding}~\emph{et al.} \cite{ding2021simple} put forward \textbf{TARGCN} \textit{(Time-Aware Relational Graph Convolutional Network)}, a time-aware relational graph encoder designed for the TKGC task. It can achieve greater expressiveness with a smaller number of parameters. %\WangC{Can we revise this sentence into `significantly reduce the number of parameters.'} 
TARGCN explores the temporal context of each entity to learn entity embeddings and models temporal differences to encode temporal information through a functional temporal encoder. %The aforementioned methods encounter challenges related to scalability and insufficient global information. To address the problem of insufficient global information, 
\textbf{TASTER} \cite{wang2023temporal} explores the evolution process of entities via a sparse transformation matrix and simultaneously models the local information in a specific timestamp and global information. %\WangC{What is the meaning of plausibility? I believe this sentence is meaningless.} %In addition, TASTER explores the evolution process of entities via a sparse transformation matrix. 
Subsequently, TASTER respectively models entity association and evolution to overcome scalability limitations. \textbf{Time-LowFER} \cite{dikeoulias2022temporal} proposes a cycle-aware time-encoding function to decompose the timestamp into four important components including \textit{year}, \textit{month}, \textit{week} and \textit{day}, so as to better encode the timestamp. Afterwards, Time-LowFER models the association of TKG through LowFER \cite{amin2020lowfer}, which introduces a low-rank tensor decomposition mechanism to facilitate the interaction between entities and relations.

%\subsubsection{Diachronic Embedding Function-based TKGC Methods}
Moreover, special transformation functions, i.e., diachronic embedding functions, can encode timestamps or associate timestamps with entities and relations more efficiently \cite{goel2020diachronic, wang2022dynamic}. 
\textbf{Goel}~\emph{et al.} \cite{goel2020diachronic} first propose a general diachronic embedding function (such as DE-SimplE and DE-DistMult), which is model-independent and can be combined with any SKGC method. The diachronic embedding function can obtain the entity representation at any timestamp. 
Although DE-SimplE is capable of capturing temporal information of TKGs, it is insufficient to exploit the complex structure of the graph and its entity modeling is not comprehensive enough. 
\textbf{DEGAT} \cite{wang2022dynamic} exploits the \textit{Graph ATtention network (GAT)} to capture the complex graph structure and introduces the diachronic embedding function to model the association of the entity and timestamp. 

\begin{figure}[t]
    \begin{center}
    \includegraphics[scale=0.56]{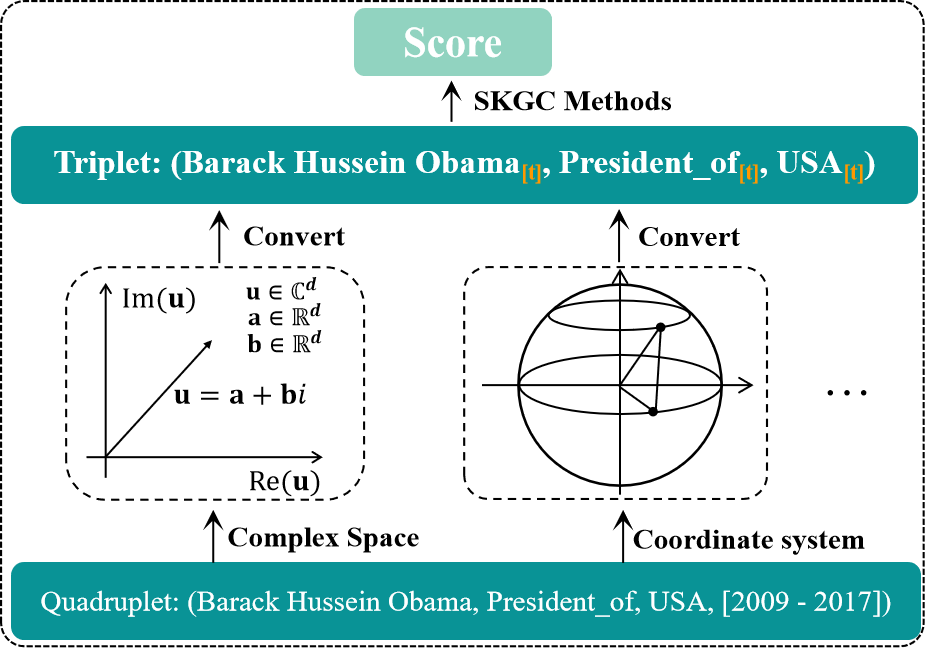}
    \end{center}
     \caption{The framework of the complex embedding functions-based TKGCs.}
    \label{fig:my_label_CE}
\end{figure}

\subsubsection{Complex Embedding Functions}
Complex embedding functions-based TKGC methods typically embed TKGs into complex spaces or special coordinate systems to capture various relational patterns, such as symmetry (e.g., spouse), antisymmetry (e.g., predecessor), inversion (e.g., hypernym and hyponym) and composition (e.g., my son’s mother is my wife) \cite{DBLP:conf/kdd/XiongZNXP0S22,DBLP:journals/corr/abs-2306-02199,Sun2019RotatEKG}, and semantic information. 
%This is because it enables generalization capability, i.e, once the patterns are learned, new facts that respect the patterns can be directly inferred. A key example is the KG embedding in complex or hypercomplex space \cite{Sun2019RotatEKG}. This has been generalized to TKGC methods, which embed TKG into the complex or quaternion space to model and infer various relation patterns. Then, they associate timestamps with entities or relations in complex or quaternion space and obtain the score of the quadruplet through any SKGC methods. 
As shown in Fig. \ref{fig:my_label_CE}, the quadruplet \textit{(Barack Hussein Obama, President\_of, USA, [2009 - 2017])} is converted into the triplet by associating timestamps with entities or relations in complex space or polar/spherical coordinate system, and then the missing item can be predicted through SKGC methods. 

\textbf{ChronoR} \cite{sadeghian2021chronor} is an extension of RotatE \cite{Sun2019RotatEKG} that embeds the triplet $(\mathbf{s}, \mathbf{r}, \mathbf{o})$ into the complex space and interprets the relation as a rotation from the head entity to the tail entity. 
Similarly, ChronoR associates the timestamp with the relation and treats the combination of relation and timestamp as a rotation from the head entity to the tail entity. The score function is defined as follows,
\begin{equation}
    g(\mathbf{s},\mathbf{r},\mathbf{o},\mathbf{t})=\langle \mathbf{s} \circ [\mathbf{r}|\mathbf{t}] \circ \mathbf{r}_2, \mathbf{o} \rangle,
\end{equation}
where $\mathbf{s},\ \mathbf{r},\ \mathbf{o},\ \mathbf{t} \in \mathbb{C}^{d}$; $\circ$ denotes Hadamard (or element-wise) product; $\langle \cdot \rangle$ denotes the inner product operation.

\textbf{TComplEx} and \textbf{TNTComplEx} \cite{lacroix2020tensor} initially expand the $3$rd-order tensor to the $4$th-order in complex space %\WangC{How about $4$th-order complex space?} 
to perform TKGC. 
Especially, TNTComplEx considers that some facts may not change over time,  thus dividing the TKG into temporal and non-temporal components. 
Equally important, \textbf{TeRo} \cite{xu2020tero} incorporates timestamps into the head and tail entities in complex space to define the temporal evolution of entities, and the relation is served as the rotation from the head to the tail entity. 
To further improve the modeling and reasoning ability of the temporal relation pattern, \textbf{TGeomE} \cite{9713947} embeds the TKG into the hypercomplex (quaternion) space, and incorporates the timestamp into relation to define a time-specific relation embedding. 
Meanwhile, TGeomE performs geometric product and Clifford conjugation operations among the head entity, the time-specific relation and the tail entity to denote the score function in quaternion space. Based on TGeomE, \textbf{TeLM} \cite{xu2021temporal} further proposes a novel temporal regularization for temporal embeddings to improve smoothness between the neighboring timestamps. 
\textbf{RotateQVS} \cite{chen2022rotateqvs} regards temporal information as the rotation axis and employs a rotation on the entity to represent the evolution of the entity in the quaternion space. 
In order to comprehensively capture the spatio-temporal and relational dependencies in TKG, \textbf{ST-NewDE} \cite{nayyeri2022dihedron} encodes the TKG into a rich geometric space and uses Dihedron algebra to learn such spatial and temporal aspects. Recently, \textbf{BiQCap} \cite{zhang2023biqcap} explores the evolution of the entity and represents each temporal entity as a translation, while each relation is represented as a combination of Euclidean rotation and hyperbolic rotation in biquaternion space.

The aforementioned methods embed TKGs into complex or quaternion spaces, which achieve the capturing of complex relation patterns%, such as symmetry, antisymmetry, inversion and composition
. However, they face challenges in capturing the semantic information within TKGs. Currently, some researchers attempt to embed TKGs into special coordinate systems to uncover their semantic information.
\textbf{HA-TKGE} \cite{10030053} divides temporal information into three hierarchies: year \textit{(Y)}, month \textit{(Y-M)}, and day \textit{(Y-M-D)} (semantic hierarchy in descending order: year, month, day). It hierarchically encodes temporal information into a polar coordinate system to fully exploit the semantic information in TKG. Specifically, HA-TKGE uses radial coordinates to represent temporal information at different levels, where entities with smaller radii indicate a higher semantic hierarchy. 
Angular coordinates model temporal information at the same semantic hierarchy. Similarly, \textbf{STKE} \cite{wang2022stke} embeds TKG into a spherical coordinate system, and regards each fact as a rotation from the head entity to the tail entity. Specifically, STKE divides each fact into the radial part, the azimuth part, and the polar part to learn the accurate embeddings of each quadruplet. However, STKE cannot model and infer complex relation patterns. \textbf{HTKE} \cite{he2022hyperplane} simultaneously embeds the knowledge that happen at the same timestamp into both a polar coordinate system and a temporal hyperplane to model complex relation patterns.

\begin{figure}[t]
    \begin{center}
    \includegraphics[scale=0.58]{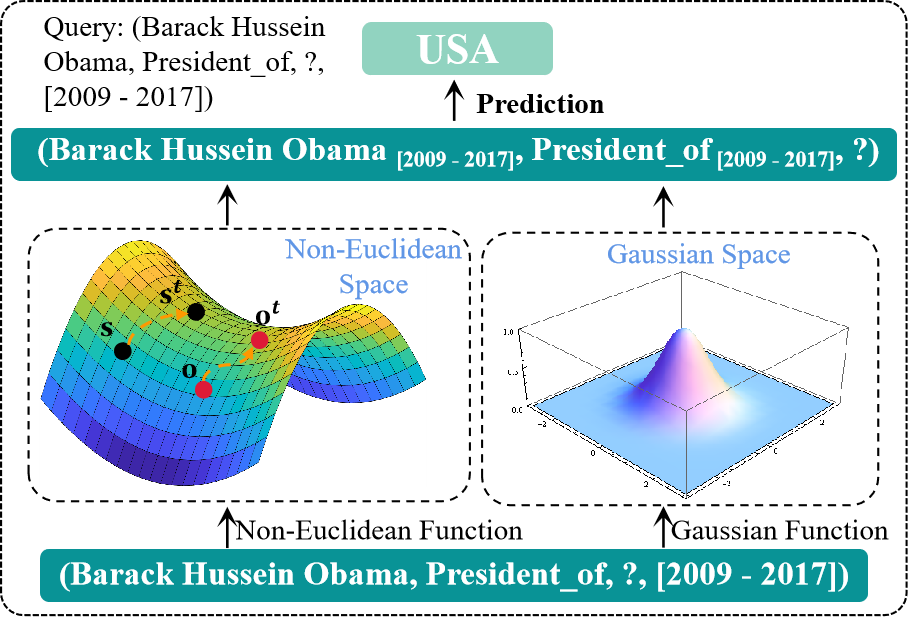}
    \end{center}
    \caption{The framework of the non-linear embedding functions-based TKGC methods, which embeds the TKG through \textit{Non-Euclidean Function} and \textit{Gaussian Function} to capture the temporal uncertainty, the semantic and structural information hidden in the TKG.}
    \label{fig:my_label_ST}
\end{figure}

\subsubsection{Non-linear Embedding Functions}
Non-linear embedding functions-based TKGC methods typically utilize non-linear functions, such as Gaussian and non-Euclidean functions, to embed TKGs, so as to deeply capture the temporal uncertainty, the semantic and structural information. 
The framework of the non-linear embedding functions-based TKGC methods is shown in Fig. \ref{fig:my_label_ST}. 
Here, we summarize some classic non-linear embedding functions-based TKGC methods, such as \textbf{DyERNIE} \cite{han2020dyernie}, \textbf{ATiSE} \cite{xu2020temporal} and \textbf{HTKE} \cite{he2022hyperplane}.

\textbf{DyERNIE} \cite{han2020dyernie} proposes a non-Euclidean embedding function that explores evolving entity representations through a velocity vector defined in the tangent space at each timestamp. 
Specifically, DyERNIE embeds the TKG into the Riemannian manifold and introduces an entity-specific velocity vector to capture dynamic facts that change over time. The evolution process can be denoted as,
\begin{equation}
    \mathbf{s}(t) = \text{exp}_{\mathbf{0}}^c(\text{log}_{\mathbf{0}}^c(\mathbf{\bar{s}})+\mathbf{v}_{\mathbf{s}}t)
\end{equation}
where $\mathbf{\bar{s}}\in\mathcal{M}_c^D$ represents the entity embedding that does not change over time in manifold space with the curvature $c$ and
the dimension $D$; $\mathbf{v}_{\mathbf{s}}\in\mathcal{T}_{\mathbf{0}}\mathcal{M}_c^D$ represents an entity-specific velocity vector that is defined in the tangent space at origin $\mathbf{0}$ and captures evolution of the entity $\mathbf{s}$ over time. 
Moreover, DyERNIE measures the distance \cite{skopek2020mixed} between the head entity $\mathbf{s}(t)$ and the tail entity $\mathbf{o}(t)$ to define the score function. Likewise, \textbf{HERCULES} \cite{montella2021hyperbolic} is a time-aware extension of ATTH \cite{2020Low}, which embeds TKG into the hyperbolic space to fully model different relation patterns and hierarchical structure of TKG. Equally important, HERCULES defines the curvature of a Riemannian manifold as the product of the relation and temporal information, which captures the evolution process of relation.

\textbf{ATiSE} \cite{xu2020temporal} initially embeds TKG into the space of multi-dimensional Gaussian distributions and regards the evolution of the entity/relation representation as an additive time series, comprising the trend component, seasonal component, and random component. ATiSE considers the temporal uncertainty during the evolution of entity/relation representations over time. Similarly, \textbf{TKGC-AGP} \cite{zhang2022temporal} embeds entities and relations of TKG into a specific \textit{Multivariate Gaussian Process (MGP)} to improve the flexibility and expressiveness of TKG, and models the temporal uncertainty through the kernel function and entity/relation-specific covariance matrix of MGP. Furthermore, a $1$st-order Markov assumption-based algorithm is designed to effectively optimize the training process of TKGC-AGP. 

\subsection{Deep Learning-based TKGCs}
\begin{figure}[t]
    \begin{center}
    \includegraphics[scale=0.6]{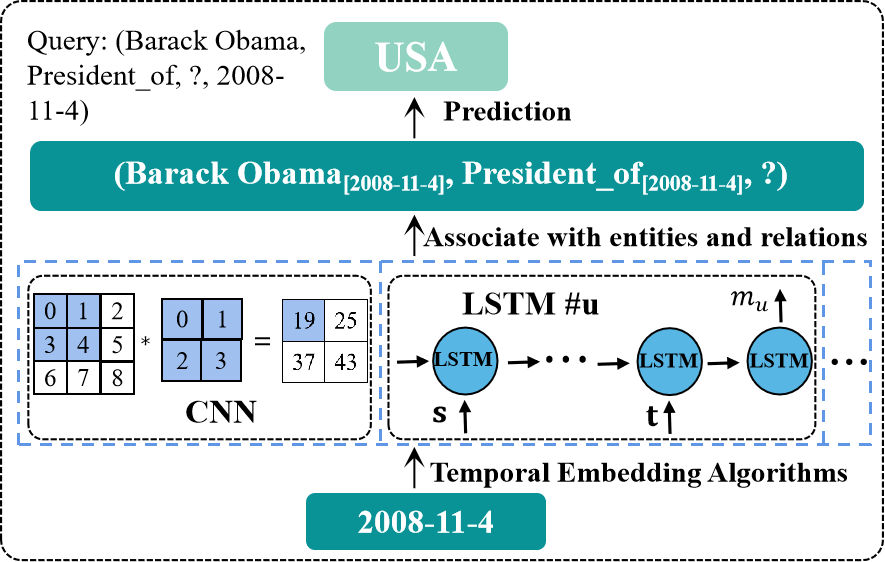}
    \end{center}
    \caption{The framework of the deep learning-based TKGC methods.}
    \label{fig:my_label_DP}
\end{figure}

%Deep learning-based TKGC methods aim to utilize deep learning techniques to process the temporal information in TKG.
Thanks to the powerful information mining ability, various deep learning algorithms are used to process temporal information in TKGs. 
As depicted in Fig. \ref{fig:my_label_DP}, deep learning-based TKGC methods employ
%deep learning techniques, such as 
CNN or LSTM to encode the timestamp ``\textit{2008-11-4}". 
%This enables the extraction of temporal representations, allowing for the exploration of the entity and relation evolution through capturing the intrinsic correlations. 
Then, the encoded timestamp supports the entity and relation evolution by capturing their intrinsic correlations.

\subsubsection{Timestamps-Specific Space}
Temporal-specific space-based TKGC methods \cite{dasgupta2018hyte, he2022hyperplane, wang2023qdn} generally encode timestamps as specific hyperplanes or semantic spaces to enhance the expression of temporal information.

\textbf{HyTE} \cite{dasgupta2018hyte} associates each timestamp with a corresponding hyperplane, and then maps entities and relations into this hyperplane to perform the translation operation. However, it is worth noting that HyTE is limited in its ability to model and infer complex relation patterns. %, such as symmetry, antisymmetry, inversion, and composition. 
\textbf{HTKE} \cite{he2022hyperplane} embeds knowledge that happened at the same timestamp into a polar coordinate system, effectively capturing complex relation patterns. It introduces time-specific hyperplanes to explicitly incorporate temporal information with entities and relations. 

However, both HyTE and HTKE ignore the diversity of potential temporal properties and relations, as well as the temporal dependency between neighboring hyperplanes. Consequently, \textbf{TRHyTE} \cite{yuan2022trhyte} defines three typical temporal properties, such as time interval, open interval, and time point, to distinguish different situations. Specifically, TRHyTE sequentially maps entities to the relational space and subsequently to the temporal hyperplane, enabling the learning of time-relation-aware embeddings. Additionally, TRHyTE applies \textit{Gate Recurrent Unit (GRU)} to simulate the evolution process, so as to capture the temporal dependency between neighboring hyperplanes.  Nevertheless, TRHyTE ignores the balance of timestamp distribution, which will severely limit the expressiveness of the TKGC method. \textbf{BTHyTE} \cite{liu2021temporal} proposes the HyTE \cite{dasgupta2018hyte}-based direct encoding temporal information model to embed the timestamp and sets the finest granularity to ensure a balanced distribution of the number of facts in each finest granularity cell. %Subsequently, BTHyTE leverages HyTE \cite{dasgupta2018hyte} to accomplish the TKGC task.

\begin{table}[h]
    \centering
    \caption{Representation of the timestamp.}
    \setlength{\tabcolsep}{1.0mm}{
    \begin{tabular}{|c|c c c c| c c c| c c c c c| c c c c c c c|}
    \hline
    \rule{0pt}{9pt}
    {Section}&\multicolumn{4}{c|}{Quarters}&\multicolumn{3}{c|}{Months}&\multicolumn{5}{c|}{Weeks}&\multicolumn{7}{c|}{Days}\\
    \hline
    \rule{0pt}{9pt}
    {12 June, 2023}&{0}&{1}&{0}&{0}&{0}&{0}&{1}&{0}&{1}&{0}&{0}&{0}&{0}&{0}&{0}&{0}&{1}&{0}&{0}\\
   \hline
   \end{tabular}}
\label{tab:label_DL}
\end{table}

\textbf{ToKEi} \cite{leblay2020towards} is also based on the HyTE, representing the timestamp within a vector of four sections of sizes $4$, $3$, $5$, and $7$, respectively. Specifically, the first section represents \textit{Quarters}, the second section denotes \textit{Months}, the next section is \textit{Weeks}, and \textit{Days} is the final section. More precisely, a year consists of four quarters, a quarter contains three months, a month encompasses five weeks, and a week comprises seven days. As shown in TABLE \ref{tab:label_DL}, 
ToKEi sets the second position of the \textit{Quarters} to
$1$ (Due to June is in the second quarter) and the rest to $0$. Similarly, the third position of the \textit{Months} (corresponding to June being the third month of the second quarter), the second position of the \textit{Weeks} (as the 12th of June falls in the second week) and the fifth position of the \textit{Days}, are set to $1$. According to the above operations, ToKEi can obtain the embedding of timestamps. Finally, ToKEi associates timestamps with entities and relations to explore the temporal evolution and accomplish the missing items through HyTE.
%\subsubsection{Temporal-specific semantic space-based TKGC methods}
%Temporal-specific semantic space-based TKGC methods \cite{li2022each, wang2023qdn} typically embed the temporal information into the temporal-specific semantic space to fully capture its semantic information.
\textbf{SANe} \cite{li2022each} employs an adaptive approach to learn different latent spaces for temporal snapshots at different timestamps and introduces \textit{Convolutional Neural Networks (CNN)} to embed KGs of different timestamps into their respective latent spaces. Moreover, SANe assigns different CNN-specific parameters for different timestamps to address the problem of overlapping latent spaces. 

However, the above methods are all modeled based on triplets, i.e., they associate timestamps with entities and relations to translate quadruplets into triplets. Undoubtedly, this severely limits the ability to express temporal information. In contrast, \textbf{QDN} \cite{wang2023qdn} is proposed as an extension of TDN \cite{wang2023tdn}, independently handling timestamps, entities, and relations in their respective spaces to comprehensively capture their semantics. Afterwards, QDN creatively designs the \textit{Quadruplet Distributor (QD)} to facilitate  the representation learning of TKG through the information aggregation and distribution among timestamps, relations and entities. In addition, QDN extends the $3$rd-order tensor into $4$th-order to build the intrinsic correlation of entities, relations, and timestamps.

\subsubsection{Long Short-Term Memory}
\textit{Long Short-Term Memory (LSTM)}-based TKGC methods \cite{tang2020timespan, li2023embedding} encode timestamps and the evolution of events by \textit{Recurrent Neural Network (RNN)} \cite{medsker2001recurrent}, LSTM \cite{1997Long} or its variants \cite{Cho2014OnTP}.

\textbf{TA-TransE} and \textbf{TA-DistMult} \cite{Learning2018} first decompose the timestamp into a series of temporal tokens, such as \textit{year}, \textit{month} and \textit{day}. For example, given a fact \textit{(Barack Hussein Obama, born in, USA, 1961-8-4)}, the timestamp ``\textit{1961-8-4}" can be represented as \textit{[1y, 9y, 6y, 1y, 8m, 4d]}. Afterward, TA-TransE and TA-DistMult apply RNNs to learn the relation representation incorporated with the timestamp. Finally, they predict the missing item of the quadruplet through TransE \cite{bordes2013translating} and DistMult \cite{2014Embedding}, respectively. 
%However, TransE and DistMult exhibit the weak expressive power. 
\textbf{TDG2E} \cite{tang2020timespan} applies the \textit{Gated Recurrent Unit (GRU)} to capture the structural information of each semantic KG, while preserving the evolution process of the TKG. In addition, TDG2E designs a timespan gate within GRU to solve the problem of unbalanced timestamp distribution in TKG. Specifically, the timespan gate can effectively associate the timestamp between the neighboring static KGs. Likewise, \textbf{TRHyTE} \cite{yuan2022trhyte} sequentially maps entities to the relational space and subsequently to the temporal hyperplane, and employs GRU to simulate the evolution process, so as to capture the temporal dependency between neighboring hyperplanes.

%However, TDG2E models all facts of TKG in a unified space and ignores the uncertain information. 
\textbf{Ma}~\emph{et al.} \cite{ma2021learning} decompose the timestamp into sequences through the bag of words model and introduce \textit{Bi-directional Long Short-Term Memory (Bi-LSTM)} to capture the semantic properties of the timestamp together with relations and entities.
To further enhance the utilization of uncertain information in TKG,
\textbf{CTRIEJ} \cite{li2023embedding} employs the GRU-based sequence model to integrate the uncertainty, structural and temporal information. Equally important, CTRIEJ introduces the self-adversarial negative sampling technique to generate negative samples, so as to improve the model expression ability.
Similarly, \textbf{TeCre} \cite{ma2023tecre} vectorizes the timestamp by representing it as a sequence of \textit{year}, \textit{month} and \textit{day}. Moreover, in order to ensure that the model is valid enough for large and complex datasets, TeCre trains the \textit{Long Short-Term Memory (LSTM)} network to learn the joint representation of timestamps and relations. Finally, it creatively designs a novel loss function to guarantee the consistency between entities and relations.

%\textbf{TeCre} \cite{ma2023tecre} trains the LSTM network to learn the evolution of relations, and then creatively designs a novel loss function to guarantee the consistency between entities and relations.

\subsubsection{Temporal Constraint}
Temporal constraint-based TKGC methods \cite{chekol2017marrying, wang2019novel} usually regard temporal information as the constraint to ensure the reasoning path along the right direction.

\textbf{Chekol}~\emph{et al.} \cite{chekol2017marrying} simultaneously capture the uncertainty and temporality of TKG  and explore \textit{Markov Logic Networks (MLNs)} and \textit{Probabilistic Soft Logics (PSLs)} to achieve reasoning tasks on TKG. However, this method has high computational complexity and limited running efficiency. In contrast, \textbf{Kgedl} \cite{wang2019novel}  captures the evolution process of TKG by embedding entities, relations and path structures. In particular, Kgedl further applies temporal information to constrain path reasoning and representation learning of entities. Likewise, \textbf{T-GAP} \cite{jung2021learning} introduces the novel temporal \textit{Graph Neural Network (GNN)} to adaptively aggregate the query-specific sub-TKG, and further encodes the temporal displacement between the timestamp of query and each edge. Furthermore, T-GAP performs a path-based inference operation over the sub-TKG to reason out the missing item. \textbf{TempCaps} \cite{fu2022tempcaps} is a light-weighted capsule network \cite{vu2019capsule}-based embedding method, which dynamically routes retrieved relations and entities in TKG. 
Specifically, TempCaps mainly consists of two important components, including the neighbor selector and the dynamic routing aggregator. More specifically, the neighbor selector is imposed with temporal constraints to facilitate the selection process. Based on the results of the neighbor selector, the dynamic routing aggregator further reasons and aggregates the neighbors, so as to dynamically learn the contextualized embedding of the query. %\textbf{IMR} \cite{du2023imf} transforms path-based reasoning into question answering and defines three indicators, including the query matching degree, answer completion level, and path confidence, for multi-hop reasoning. Additionally, IMR creatively develops an algorithm to integrate paths with different hops using the same criteria.

%\subsubsection{Temporal attention-based TKGC methods}
%Temporal attention-based TKGC methods \cite{bai2023roan, nie2023temporal} aim to impose the attention to the temporal information to enhance its importance.

    \begin{figure}[t]
    \begin{center}
    \includegraphics[scale=0.55]{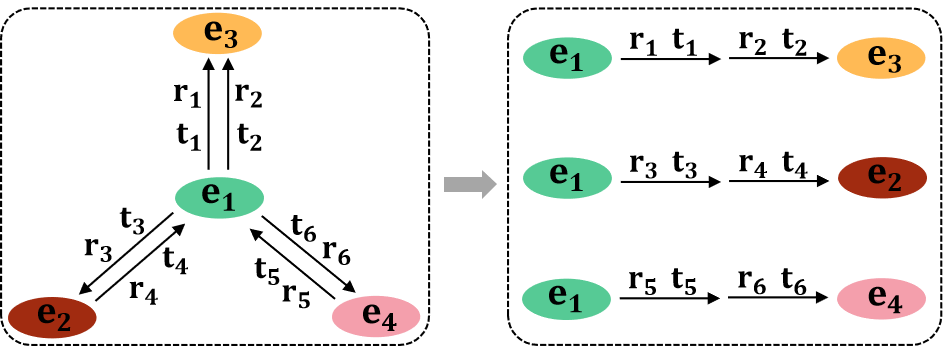}
    \end{center}
    \caption{The translation of multi-edges mesh form of TKG to the relational multi-chains forms of TKG.}
    \label{fig:my_label_AT}
\end{figure}

Existing attention-based methods primarily emphasize entity learning and may even update entities through the original embeddings of relations. Consequently, the importance of relations is significantly diminished in these methods. \textbf{RoAN} \cite{bai2023roan} proposes a relation-oriented attention mechanism that enhances the impact of relations. Specifically, RoAN reconstructs the multi-edge mesh form of TKG as the relational multi-chain form of TKG (shown in Fig. \ref{fig:my_label_AT}). Afterwards, RoAN adaptively assigns different weights to different relations to achieve the optimization process of relations. Furthermore, while RoAN focuses on the importance of relations, it ignores the importance of temporal information as well as structural information in TKG. 
\textbf{TAL-TKGC} \cite{nie2023temporal} designs a temporal attention module that captures the intrinsic correlation between timestamps and entities, and introduces the weighted GCN module to explore the structure information of the entire TKG.

\begin{table*}[t]
\centering
\caption{Summary of the interpolation TKGC methods.}
\begin{threeparttable}
  \setlength{\tabcolsep}{0.4mm}{
  \begin{tabular}{c c c c|c c c c}
  \hline
  \rule{0pt}{9pt}
  {Methods}&{Category}&{Technique}&{SKGC}&{Methods}&{Category}&{Technique}&{SKGC}\\
  \hline
  \rule{0pt}{9pt}
  {TTransE(2018)\cite{leblay2018deriving}}&{Timestamps Dep}&{Translation}&{\checkmark}&{ChronoR(2021) \cite{sadeghian2021chronor}}&{Complex Emb}&{Complex Space}&{\checkmark}\\
  \rule{0pt}{9pt}
  {TuckERTNT(2022) \cite{shao2022tucker}}&{Timestamps Dep}&{Tensor Decomp}&{\checkmark}&{TeLM(2021) \cite{xu2021temporal}}&{Complex Emb}&{Quaternion Space}&{\checkmark}\\
  \rule{0pt}{9pt}
  {ST-TransE(2020) \cite{9450214}}&{Timestamps Dep}&{Translation}&{\checkmark}&{RotateQVS(2022) \cite{chen2022rotateqvs}}&{Complex Emb}&{Quaternion Space}&{\checkmark}\\
  \rule{0pt}{9pt}
  {T-SimplE(2020) \cite{9288194}}&{Timestamps Dep}&{Tensor Decomp}&{\checkmark}&{ST-NewDE(2022) \cite{nayyeri2022dihedron}}&{Complex Emb}&{Dihedron Algebra}&{\checkmark}\\
  \rule{0pt}{9pt}
  {TKGFrame(2020) \cite{zhang2020tkgframe}}&{Timestamps Dep}&{ILP}&{--}&{TGeomE(2023) \cite{9713947}}&{Complex Emb}&{Quaternion Space}&{\checkmark}\\
  \rule{0pt}{9pt}
  {HERCULES(2021) \cite{montella2021hyperbolic}}&{Timestamps Dep}&{Manifold}&{\checkmark}&{BiQCap(2023) \cite{zhang2023biqcap}}&{Complex Emb}&{Biquaternions/Manifold}&{\checkmark}\\
  \rule{0pt}{9pt}
  {TBDRI(2023) \cite{yu2023tbdri}}&{Timestamps Dep}&{Block Decomp}&{\checkmark}& {TA-DistMult(2018) \cite{Learning2018}}&{Decomp-Time}&{Time-Encoding}&{\checkmark}\\
  \rule{0pt}{9pt}
  {TransR-ILP(2016) \cite{jiang2016towards}}&{Transform Func}&{Translation}&{\checkmark}&{TeCre(2018) \cite{ma2023tecre}}&{Decomp-Time}&{Time-Encoding/LSTM}&{\checkmark}\\
  \rule{0pt}{9pt}
  {DE-SimplE(2020) \cite{goel2020diachronic}}&{Transform Func}&{Tensor Decomp}&{\checkmark}&{ToKEi(2020) \cite{leblay2020towards}}&{Decomp-Time}&{Time-Encoding}&{\checkmark}\\
  \rule{0pt}{9pt}
  {TARGCN(2021) \cite{ding2021simple}}&{Transform Func}&{GCN}&{\checkmark}&{LBiE(2021) \cite{ma2021learning}}&{Decomp-Time}&{Bag of Words/BiLSTM}&{\checkmark}\\
  \rule{0pt}{9pt}
  {DEGAT(2022) \cite{wang2022dynamic}}&{Transform Func}&{GAT}&{--}&{Time-LowFER(2022) \cite{dikeoulias2022temporal}}&{Decomp-Time}&{Tensor Decomp}&{\checkmark}\\
  \rule{0pt}{9pt}
  {TNTSimplE(2022) \cite{he2022improving}}&{Transform Func}&{CP Decomp}&{\checkmark}&{HyTE(2018) \cite{dasgupta2018hyte}}&{Temporal-Specific}&{Temporal-Hyperp}&{\checkmark}\\
  \rule{0pt}{9pt}
  {BoxTE(2022) \cite{messner2022temporal}}&{Transform Func}&{Translation}&{\checkmark}&{TRHyTE(2021) \cite{yuan2022trhyte}}&{Temporal-Specific}&{Temporal-Hyperp/GRU}&{\checkmark}\\
  \rule{0pt}{9pt}
  {F-BoxTE(2022) \cite{dai2022wasserstein}}&{Transform Func}&{Adversarial Learn}&{\checkmark}&{BTHyTE(2021) \cite{liu2021temporal}}&{Temporal-Specific}&{Temporal-Hyperp}&{\checkmark}\\
  \rule{0pt}{9pt}
  {SPLIME(2023) \cite{radstok2021leveraging}}&{Transform Func}&{Transformation}&{\checkmark}&{SANe(2022) \cite{li2022each}}&{Temporal-Specific}&{Multi-semantic Space}&{--}\\
  \rule{0pt}{9pt}
  {TASTER(2023) \cite{wang2023temporal}}&{Transform Func}&{Sparse Matrix}&{\checkmark}&{QDN(2023) \cite{wang2023qdn}}&{Temporal-Specific}&{Tensor Decomposition}&{--}\\
  \rule{0pt}{9pt}
  {HTTR(2023) \cite{zhao2023householder}}&{Transform Func}&{Householder}&{\checkmark}&{TDG2E(2020) \cite{tang2020timespan}}&{LSTM}&{GRU}&{--}\\
  \rule{0pt}{9pt}
  {DyERNIE(2020) \cite{han2020dyernie}}&{Special Space}&{Manifold}&{\checkmark}&{CTRIEJ(2023) \cite{li2023embedding}}&{LSTM}&{GRU/Negative Sampling}&{\checkmark}\\
  \rule{0pt}{9pt}
  {ATiSE(2020) \cite{xu2020temporal}}&{Special Space}&{Gaussian}&{--}&{MUT(2017) \cite{chekol2017marrying}}&{Constraint}&{Markov}&{--}\\
  \rule{0pt}{9pt}
  {TKGC-AGP(2022) \cite{zhang2022temporal}}&{Special Space}&{Gaussian/Markov}&{--}&{Kgedl(2019) \cite{wang2019novel}}&{Constraint}&{Path Reasoning}&{--}\\
  \rule{0pt}{9pt}
  {HA-TKGE(2022) \cite{10030053}}&{Special Space}&{Polar CS}&{\checkmark}&{T-GAP(2021) \cite{jung2021learning}}&{Constraint}&{GNN}&{--}\\
  \rule{0pt}{9pt}
  {HTKE(2022) \cite{he2022hyperplane}}&{Special Space}&{Polar CS}&{\checkmark}&{TempCaps(2022) \cite{fu2022tempcaps}}&{Constraint}&{Capsule Network}&{--}\\
  \rule{0pt}{9pt}
  {STKE(2023) \cite{wang2022stke}}&{Special Space}&{Spherical CS}&{\checkmark}&{IMR(2023) \cite{du2023imf}}&{Constraint}&{Path Reasoning}&{--}\\
  \rule{0pt}{9pt}
  {TNTComplEx(2020) \cite{lacroix2020tensor}}&{Complex Emb}&{Complex Space}&{\checkmark}&{RoAN(2023) \cite{bai2023roan}}&{Constraint}&{Attention}&{\checkmark}\\
  \rule{0pt}{9pt}
  {TeRo(2020) \cite{xu2020tero}}&{Complex Emb}&{Complex Space}&{\checkmark}&{TAL-TKGC(2023) \cite{nie2023temporal}}&{Constraint}&{Attention}&{\checkmark}\\
\hline
\end{tabular}
\begin{tablenotes}
        \footnotesize              
        \item[1] \textbf{SKGC} represents  whether TKGC methods are modeled based on SKGC methods, and draws \checkmark if it is;
        \item[2] %\textbf{Transform Func} denotes \textit{Transformation function-based TKGC methods}; 
        \textbf{Constraint} is \textit{Temporal constraint-based TKGC methods}. %\textbf{Decomp-Time} is \textit{Decomposed temporal information-based TKGC methods}.
        %\textbf{GCN} denotes \textit{Graph convolutional neural network}; \textbf{GAT} represents \textit{Graph Attention Network}; 
        \textbf{Temporal-Hyperp} is \textit{Temporal Hyperplane}; %\textbf{TD/BD/CD} represent \textit{Tensor/Block/Canonical-polyadic Decomposition}, respectively; 
        %\textbf{Householder} is \textit{Householder transformation}; 
        \textbf{ILP} represents \textit{Integer Linear Programming}; \textbf{CS} denote \textit{Coordinate System}. 
      \end{tablenotes}
}
\end{threeparttable}
\label{tab:label_inter}
\end{table*}

\section{Extrapolation-based TKGCs}\label{Sec:Extrapolation}

In this section, we group the extrapolation-based TKGC methods into the following several categories: \textit{Rule-based TKGC methods}, \textit{Graph neural network-based TKGC methods}, \textit{Reinforcement learning-based TKGC methods} and \textit{Meta learning-based TKGC methods}.

\subsection{Rule-based TKGCs}

\begin{figure}[t]
    \begin{center}
    \includegraphics[width=1\columnwidth]{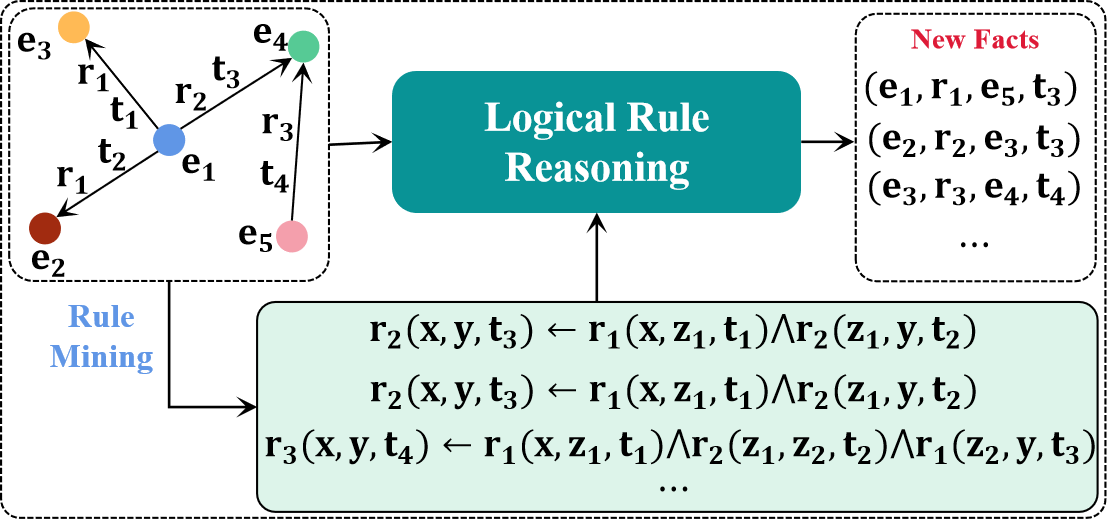}
    \end{center}
    \caption{The framework of the rule-based TKGC methods.}
    \label{fig:rule_based}
\end{figure}

Rule-based TKGC methods are often praised due to their interpretability and reliability, which obtains great success in static knowledge graph applications \cite{yang2017differentiable,sadeghian2019drum}. 
Recently, researchers explore the potential that applies the rule-based methods for TKGC. 
As shown in Fig. \ref{fig:rule_based}, rule-based TKGC methods %first try to 
extract a series of temporal logical rules from the given TKG before the reasoning operation. 
Temporal logical rules define the relationship between two entities $\mathbf{x}$ and $\mathbf{y}$ at timestamp $\mathbf{t}_l$,   
\begin{equation}
    \mathbf{r}(\mathbf{x},\mathbf{y},\mathbf{t}_l) \leftarrow \mathbf{r}_1(\mathbf{x},\mathbf{z}_1,\mathbf{t}_1) \wedge \ldots \wedge \mathbf{r}_{l-1}(\mathbf{z}_{l-1},\mathbf{y},\mathbf{t}_{l-1}),\label{eq:t_rule}
\end{equation}
where the left-hand side denotes the rule head with relation $\mathbf{r}$ that can be induced by ($\leftarrow$) the right-hand rule body. The rule body is represented by the conjunction ($\wedge$) of a series of body relations $\mathbf{r}_*$.

Each rule is written in the form of a logical implication. 
%which states that 
If the conditions on the right-hand side (rule body) are satisfied, the statement on the left-hand (rule head) holds true. 
The extracted rules are fed into a logical rule reasoning module to infer new events on TKG, applying either forward-chain or backward-chain reasoning module \cite{salvat1996sound}.

\subsubsection{Manual Rule Mining} 

Recently, many researchers analyze the characteristic of TKG and manually design some logical rules for TKGC. 
\textbf{TPRG} \cite{bai2023multi} proposes a multi-hop TKGC method based on temporal logical rules. 
TPRG analyzes the logical connections of multi-hop paths in the TKG. It manually defines fourteen temporal logical rules. These rules capture different kinds of logical relations between entities and can be used to infer new facts. 
\textbf{KGFFP} \cite{ge2022spatio} asks human experts to define several temporal logical rules %in the TKG. These rules are applied on the TKG
for forest fire prediction. 
\textbf{TLmod} \cite{bai2023temporal} analyzes the principles of the temporal logical rule definition. It proposes a pruning strategy to obtain rules and calculate confidence scores. The rules with high confidence are selected for TKGC.
 
\subsubsection{Automatic Rule Mining} 
With the increasing relations and entities in TKG, it is burdensome to manually define the logical rules. Thus, automatic rule mining has attached increasing attention from researchers. 

\textbf{ALRE-IR} \cite{mei2022adaptive} proposes an adaptive logical rule embedding model that automatically extracts logical rules from historical data. 
ALRE-IR extracts all possible rule paths between entities. Then, it adopts GRU to encode the representation of each rule. The learned rule embeddings are used to predict missing facts. \textbf{TLogic} \cite{liu2022tlogic} introduces a novel symbolic framework based on temporal random walks in TKGs. TLogic directly learns temporal logical rules from TKG and feeds these rules into a symbolic reasoning module for prediction. TLogic offers explicit and human-readable explanations in the form of temporal logical rules that can be easily scaled to accommodate large datasets. \textbf{TILP} \cite{xiong2023tilp} presents a differentiable framework for temporal logical rule learning. TILP proposes the constrained random walk mechanism on TKG. By introducing the temporal operators, TILP enables to learn temporal logical rules from TKG without restrictions. Instead of learning simple chain-like rules as shown in Eq. \ref{eq:t_rule}, \textbf{TFLEX} \cite{lin2022tflex} proposes a temporal feature-logic embedding framework that supports complex multi-hop logical rules on TKG.

\subsection{Graph Neural Network-based TKGCs}
\textit{Graph Neural Network (GNN)}-based TKGC methods generally apply GNN to explore the intrinsic topology relevance between entities or between entities and relations in TKG, so as to obtain high-quality embeddings. 
In specific, GNN-based TKGC methods mainly consist of \textit{Graph convolutional network-based TKGC methods}, \textit{Graph attention network-based TKGC methods} and \textit{Transformer-based TKGC methods}

\subsubsection{Graph Convolutional Network}
\begin{figure}[t]
    \centering
    \includegraphics[scale=0.5]{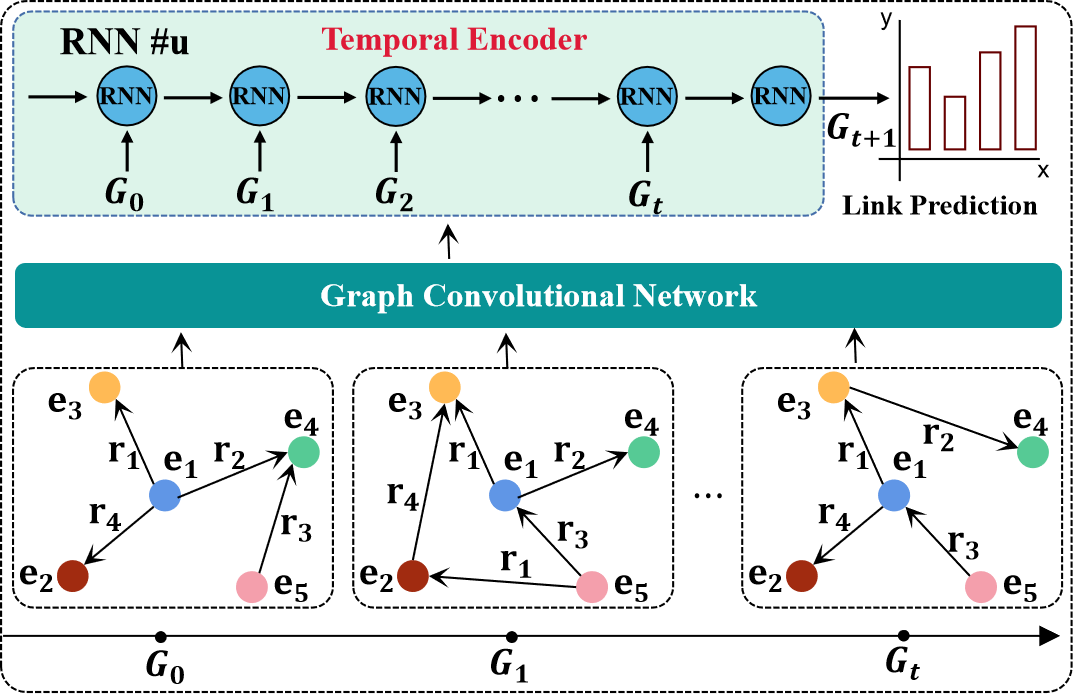}
    \caption{The framework of the graph convolutional network-based TKGC methods}
    \label{fig:gnn-tkgc}
\end{figure}

\textit{Graph Convolutional Network (GCN)}-based TKGC methods typically integrate the graph structural encoder and the temporal encoder to derive entity representations. As depicted in Fig. \ref{fig:gnn-tkgc}, each snapshot of the TKG is encoded by \textit{Graph Convolutional Network (GCN)}, while the temporal dependencies among multiple snapshots are captured by \textit{Recurrent Neural Networks (RNNs)}. 

\textbf{Structural encoder} generates entity embeddings based on the graph $G^{(t)}$ within each time step. This is typically built upon existing encoders of message passing networks on static KGs. For example, RE-NET \cite{jin2019recurrent} uses a multi-relational graph aggregator \cite{DBLP:conf/esws/SchlichtkrullKB18} to capture the graph structural information. Specifically, the multi-relational graph aggregator can incorporate information from multi-relational and multi-hop neighbors. Formally, the aggregator is defined as follows,
\begin{equation}
g\left(\mathbf{N}_t^{(\mathrm{s})}\right)=\mathbf{h}_s^{(l+1)}=\sigma(\sum_{\mathrm{r} \in \mathcal{R}} \sum_{\mathrm{s,o} \in \mathcal{E}} \frac{1}{c_s} \mathbf{W}_r^{(l)} \mathbf{h}_{\mathrm{o}}^{(l)}+\mathbf{W}_0^{(l)} \mathbf{h}_s^{(l)}),
\end{equation}
where initial hidden representation for each node $\mathbf{h}_o^{(0)}$ is set to trainable embedding vector; $\sigma$ denotes the non-linear activation function and $c_s$ is a normalizing factor. 

\textbf{Temporal Encoder} integrates temporal information into entity representations. Specifically, RE-NET \cite{jin2019recurrent} employs an RNN-based temporal encoder to capture temporal dependencies as follows,
\begin{equation}
  \begin{aligned}
\mathbf{H}_t & =\operatorname{RNN}^1\left(g\left(G_t\right), \mathbf{H}_{t-1}\right) \\
\mathbf{h}_t(\mathbf{s}, \mathbf{r}) & =\operatorname{RNN}^2(g(\mathrm{N}_t^{(\mathrm{s})}), \mathbf{H}_t, \mathbf{h}_{t-1}(\mathbf{s}, \mathbf{r})) \\
\mathbf{h}_t(\mathbf{s}) & =\operatorname{RNN}^3(g(\mathrm{N}_t^{(\mathbf{s})}), \mathbf{H}_t, \mathbf{h}_{t-1}(\mathbf{s})),
\end{aligned}  
\end{equation}
where $g$ denotes an aggregate function and $\mathrm{N}_t^{(\mathrm{s})}$ stands for all the events related to entity $\mathbf{s}$ at the current time step $t$.

Based on the above framework, subsequent studies have made improvements to the structural encoder and/or the temporal encoder by incorporating more complex message passing or recurrent neural architectures. For instance, \textbf{Deng}~\emph{et al.} \cite{DBLP:conf/kdd/DengRN20} utilize CompGCN to capture the influence of neighboring entities and event types, while incorporating GRUs to model the temporal dependency among representations.
%Another method called 
\textbf{TeMP}--\textit{Temporally Enhanced Message Passing} \cite{wu2020temp} leverages RGCN to account for the impact of neighboring entities and introduces a frequency-based gating GRU to capture the temporal dependency among inactive events.
\textbf{DACHA} \cite{chen2021dacha} proposes dual GCN to obtain entity representations, which considers information interaction on both the primal graph (i.e., entity interaction graph) and the edge graph (i.e., relation interaction graph). Moreover, a self-attentive encoder is employed to model the temporal dependency among event types.
Similarly, \textbf{RE-GCN} \cite{li2021temporal} employs RGCN to aggregate messages from neighboring entities and utilizes an auto-regressive GRU to model the temporal dependency among events. 
\textbf{CyGNet} \cite{DBLP:conf/aaai/ZhuCFCZ21} leverages a copy-generation mechanism to capture the global repetition frequency of facts. 

However, these methods are difficult to simultaneously consider the sequential, repetitive, and cyclical historical facts. 
\textbf{TiRGN} \cite{DBLP:conf/ijcai/LiS022} combines local and global historical information to capture sequential, repetitive, and cyclical patterns of historical facts. It achieves this purpose through a GNN-based encoder with double recurrent mechanism. %enabling the simultaneous evolution of entity and relation representations. 
\textbf{HiSMatch} \cite{DBLP:conf/emnlp/LiHGJP0L0GC22} integrates the background knowledge into the TKGC model via a background knowledge encoder that %. The background knowledge encoder 
is also formulated by \textbf{CompGCN} \cite{DBLP:conf/kdd/DengRN20}. HiSMatch complementally captures high-order associations among entities.
\textbf{SPA} \cite{DBLP:conf/emnlp/WangDY022} automatically designs data-specific message passing architectures for TKGC.
\textbf{TANGO} \cite{DBLP:conf/emnlp/HanDMGT21} extends neural ODE \cite{chen2018neural} to model dynamic TKGs. 
TANGO preserves the continuous nature of TKGs and encodes both temporal and structural information into continuous-time dynamic embeddings.
\textbf{HGLS} \cite{DBLP:conf/www/ZhangX0WW23} transforms the TKG sequence into a global graph to explicitly associate historical entities in different time steps.
A hierarchical RGCN module is designed to capture long-term dependencies among entities by hierarchically encoding the global graph. Besides, a gating integration module is developed to adaptively integrate long- and short-term information for each entity and relation. 
% However, these methods often face the challenge of oversmoothing, wherein the representations of neighboring entities become overly similar, due to the stacking of multiple graph convolution layers to capture the influence of distant entities.

\subsubsection{Graph Attention Network}
\begin{figure}[t]
    \begin{center}
    \includegraphics[scale=0.6]{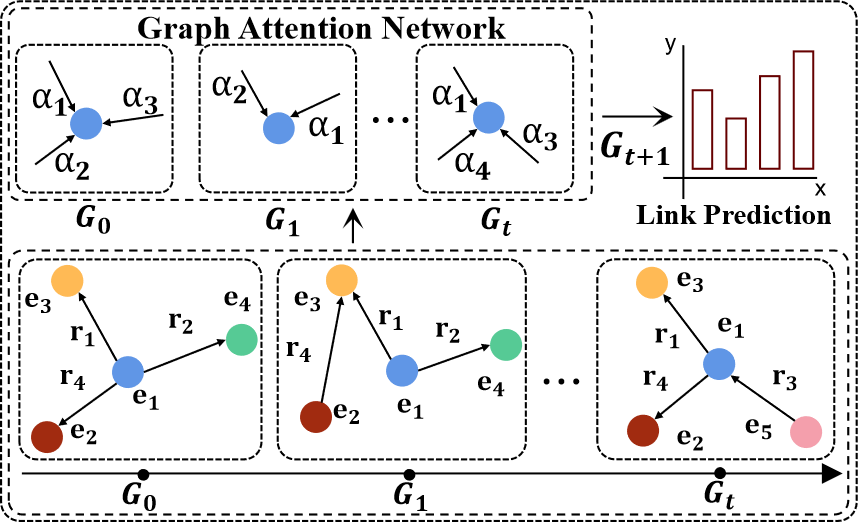}
    \end{center}
    \caption{The framework of the graph attention network-based TKGC methods.}
    \label{fig:softatt}
\end{figure}

\textit{Graph ATtention network (GAT)}-based TKGC methods %utilize the GAT mechanism to 
aggregate neighboring nodes %, thereby enhancing 
to enhance the expression capability of current node.
%The main purpose of graph attention network (GAT)-based TKGC methods is to enhance the representation ability of entities. 
%and aggregate the neighbors of entities by giving them different weights through the attention mechanism. 
%This is achieved by aggregating the neighbors of entities with different weights assigned by attention mechanism.
As illustrated in Fig. \ref{fig:softatt}, GAT-based TKGC methods %aggregate neighboring nodes by 
assign different weights to neighboring nodes to aggregate them and update the embedding of the current node to enable %, ultimately enabling
the link prediction task.

\textbf{TPmod} \cite{tpmod} defines %the concepts of 
the Goodness values for relations and the Closeness values for entity pairs, % to integrate the embeddings of entities and relations, 
which proposes a tendency strategy to fuse the values of Goodness and Closeness and utilizes the attention mechanism to aggregate historical events related to entities. %However, TPmod mainly focuses on the reasoning of SKGs and ignores the fact that every event is related to a timestamp in TKGs. 
\textbf{EvoKG} \cite{evokg} simultaneously captures the structural and temporal dynamics in TKGs by jointly modeling the event time and the evolving network structure. 
%Subsequently, EvoKG constructs interactions between entities through a neighborhood aggregation network.
However, it is crucial to note that the influence of historical events on future events is not constant and will change over time. %Different times have different impacts, 
\textbf{DA-Net} \cite{danet} learns historical information from different timestamps through an attention mechanism and allocates attention to future events.
Similarly, \textbf{TAE} \cite{tae} proposes an effective time-aware encoder that captures the impact of temporal information from entities and relations to obtain accurate time-specific representations. 
%These representations are subsequently utilized for predicting future events.
In order to explore the nature of graph evolution over time, \textbf{EvoExplore} \cite{evoexplore} describes the formation process of graph structure and the dynamic topology transformation of graphs from local and global structures, respectively. Among them, the local structure adopts the hierarchical attention mechanism to describe the establishment process of the relations. The global structure employs soft modularity parameterized by the entity representations to capture the dynamic community partition of TKGs.

Future events may occur simultaneously, and there may be mutual influences among them. \textbf{CRNet} \cite{crnet} leverages concurrent events from both history and future for TKG reasoning. 
Additionally, CRNet selects the top-$N$ candidate events and constructs a candidate graph for all missing events in the future. Subsequently, the GAT network %is used to 
handles the interaction among candidate events.
However, predicting future events not only relies on repetitive and periodic historical events but also requires the integration of potential non-historical events. 
\textbf{CENET} \cite{cenet} considers both types of information, learning a convincing distribution of entities from historical and non-historical events and identifying important entities through a comparative learning algorithm.
%There are multiple languages in the real world, but existing models often only consider one language. 
Currently, the incompleteness issue of low-resource language TKGs is particularly prominent because of the challenge in collecting sufficient corpus and annotations. 
As a consequence, this leads to suboptimal reasoning performance.
\textbf{MP-KD} \cite{mpkd} aims to enhance reasoning in low-resource TKGs by leveraging high-resource language TKGs through cross-lingual alignment and knowledge distillation.

%In order to solve the problem of multi-lingual reasoning, \textbf{MP-KD} \cite{mpkd} improves the reasoning ability of low-resource TKG by distilling knowledge from high-resource language TKG to low-resource graphs through a small set of entity alignments.

\textbf{Know-Evolve} \cite{knowEvolve} learns non-linearly evolving entity representations by considering their interactions with other entities in multi-relational space. 
It mainly models the occurrence of a fact as a multi-dimensional temporal point process, where the conditional intensity function of the process is adjusted based on the relationship score of the event. 
To consider the continuity of states in the evolution of the TKG, \textbf{RTFE} \cite{rtfe} treats the sequence of the graph as a Markov chain so that the state of the next timestamp is only related to the previous timestamp. 
When new timestamps appear, there is no need to retrain previous all timestamps, and the expansion to new timestamps occurs naturally through the state transition process. However, RTFE does not take into account recurring events. 
\textbf{CyGNet} \cite{DBLP:conf/aaai/ZhuCFCZ21} addresses this problem by learning knowledge from known events. %that have appeared in history. 
It involves predicting future events not only from the entire entity vocabulary but also by repeatedly identifying transactions and referencing known facts. 
\textbf{xERTE} \cite{han2021explainable} also designs the temporal relational graph attention mechanism to reason %guide the reasoning of 
the subgraph and poses temporal constraints to ensure the inference along the right direction. 
Likewise, \textbf{T-GAP} \cite{jung2021learning} %employs an attention mechanism to 
explores query-relevant substructure in the TKG for path-based inference. Additionally, T-GAP aggregates useful information by considering the temporal displacement between each edge and the timestamps of input query.

\begin{figure}[t]
    \begin{center}
    \includegraphics[scale=0.6]{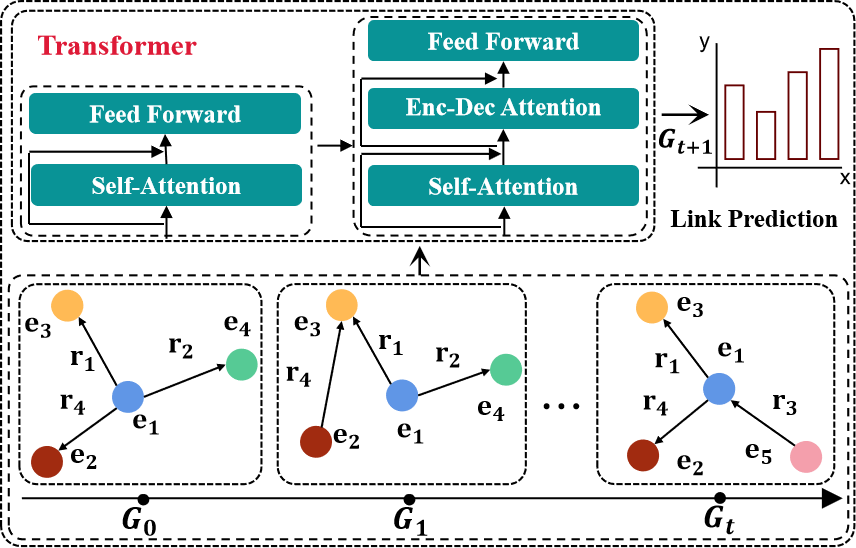}
    \end{center}
    \caption{The framework of the Transformer-based TKGC methods.}
    \label{fig:transformer}
\end{figure}

\subsubsection{Transformer} 
Transformer-based TKGC methods leverage the powerful modeling ability of Transformer to capture both structural and temporal association within TKGs. 
%In addition, they explore the internal connections and evolution trends of the TKG, so as to better infer unknown facts. 
As illustrated in Fig. \ref{fig:transformer}, Transformer not only enables the exploration of structural associations within each historical snapshot but also captures the temporal relationships among different historical snapshots to accomplish link prediction tasks. 
% using the historical snapshot and the multi-head self-attention mechanism.

Existing TKGC methods often only focus on %consider information about 
entities or relations, while ignoring the structural information of the entire TKG. \textbf{HSAE} \cite{hsae} employs a self-attention mechanism to capture the structural information of entities and relations and utilizes diachronic embedding functions to explore the evolution of entities and relations.
Events often come with certain precursors, meaning that future events often evolve from historical events. 
\textbf{rGalT} \cite{rgalt} proposes a novel auto-encoder architecture that introduces a relation-aware graph attention layer into Transformer to accommodate extrapolation inference over the TKG. %To solve the problem of not being able to predict when an event will occur, the 
\textbf{GHT} \cite{ght} captures both structural and temporal information by introducing the Transformer framework. It not only predicts the occurrence time of events but also processes unseen timestamps through a continuous-time encoding function and provides personalized query responses. 

\subsection{Meta Learning-based TKGCs}
TKGs undergo a dynamic evolutionary process with new entities and relations continuously being added. %\WangC{Please check this sentence. Why unseen?} 
These entities and relations are often unseen during the training process and are associated with only a limited number of facts, which makes it difficult for existing models to handle the future data \cite{luo2023normalizing,luo2023graph}. %\WangC{generalize the future data?}    %Meta-learning is a promising approach to address this challenge, which learns a meta-learner that can quickly adapt to new tasks with a few training examples \cite{finn2017model,nichol2018first}. 

To address this challenge, the promising meta-learning method learns a meta-learner to quickly adapt to new tasks with a few training examples \cite{finn2017model,nichol2018first}. 
In the context of TKG, the meta-learner is generalized to handle the future data with a limited number of historical facts. %\WangC{This sentence is also similar to the previous sentence.} 
The meta learning-based TKGC can be roughly grouped into two categories: \textit{matching network-based methods} and \textit{meta-optimization-based methods}.

\subsubsection{Matching Network} 
Matching network-based TKGC methods aim to learn a metric space where the distance between the few-shot historical facts and the future data can be used to predict new facts. 
\textbf{FTAG} \cite{mirtaherione} proposes a one-shot meta-learning TKGC framework, which designs %. It proposes 
a novel temporal neighborhood encoder empowered by a self-attention mechanism to capture the temporal interactions between entities and represent historical facts. Subsequently, it builds a matching network to compute the similarity score between new facts and historical examples. 
In this way, FTAG predicts new facts with only one-shot historical data. 
To handle few-shot data, \textbf{FTMF} \cite{bai2022ftmf} employs a cyclic recursive aggregation network to aggregate few-shot data and utilizes a fault-tolerant mechanism to consider the noise information. 
Finally, a RNN-based matching network is employed to measure the similarity between the few-shot data and future data. \textbf{TFSC} \cite{zhang2023few} designs a time-aware matching processor that incorporates the temporal information to calculate the similarity score. 

\subsubsection{Meta-optimization} % Gradient Optimization/ Meta-gradient
Meta-optimization-based TKGC methods update the parameters with the meta-learning objective on the few-shot samples and better generalize to future data.
%Gradient-based TKGC methods try to update the parameters of the meta-learner on the few-shot samples and better generalize to future data. 
\textbf{MOST} \cite{ding2022learning} proposes a meta-learning framework to learn meta-representation for the few-shot relations and predict new facts. 
\textbf{MetaTKG} \cite{xia2023metatkg} designs a temporal meta-learner to learn evolutionary meta-knowledge, which guides the prediction model in adapting to future data. 
Specifically, a gating integration module adaptively establishes temporal correlations between historical data. 
\textbf{MetaTKGR} \cite{wanglearning} proposes a novel meta-learning temporal knowledge graph reasoning framework. To consider the dynamic distribution shift, it dynamically adjusts the strategies of sampling and aggregating neighbors from recent facts for new entities. %These strategies are used to update the reasoning model. 
In this way, MetaTKGR enables to handle temporal adaptation with large variance. 

\begin{comment}
   \begin{figure}[t]
    \begin{center}
    \includegraphics[width=1\columnwidth]{Figures/meta_based.pdf}
    \end{center}
    \caption{The framework of the meta-learning-based TKGC methods.}
    \label{fig:meta_based}
\end{figure} 
\end{comment}

\begin{table*}[t]
\centering
\caption{Summary of the extrapolation TKGC methods.}
\begin{threeparttable}
  \setlength{\tabcolsep}{0.5mm}{
  \begin{tabular}{c c c c|c c c c}
  \hline
  \rule{0pt}{9pt}
  {Methods}&{Category}&{Technique}&{ED}&{Methods}&{Category}&{Technique}&{ED}\\
  \hline
  \rule{0pt}{9pt}
    {TPmod(2021)\cite{tpmod}}&{Graph attention}&{GRU}&{--}&{FTAG(2022)\cite{mirtaherione}}&{Meta learning}&{Attention}&{--}\\
    \rule{0pt}{9pt}
    {EvoKG(2022)\cite{evokg}}&{Graph attention}&{R-GCN}&{--}&{FTMF(2022)\cite{bai2022ftmf}}&{Meta learning}&{RNN Matching Network}&{--}\\
    \rule{0pt}{9pt}
    {DA-Net(2022)\cite{danet}}&{Graph attention}&{Attention}&{--}&{MOST(2022)\cite{ding2022learning}}&{Meta learning}&{Meta-representation Learner}&{--}\\
    \rule{0pt}{9pt}
    {TAE(2022)\cite{tae}}&{Graph attention}&{CNN}&{\checkmark}&{MetaTKGR(2022)\cite{wanglearning}}&{Meta learning}&{Temporal Domain Gen}&{--}\\
    \rule{0pt}{9pt}{EvoExplore(2022)\cite{evoexplore}}&{Graph attention}&{Attention}&{--}&{MetaTKG(2023)\cite{xia2023metatkg}}&{Meta learning}&{Gating Integration Module}&{--}\\
    \rule{0pt}{9pt}{CRNet(2022)\cite{crnet}}&{Graph attention}&{Translation}&{--}&{TFSC(2023)\cite{zhang2023few}}&{Meta learning}&{Time-aware Matching Network}&{--}\\
    \rule{0pt}{9pt}{CENET(2022)\cite{cenet}}&{Graph attention}&{Contrastive Learning}&{--}&{RE-NET(2019)\cite{jin2019recurrent}}&{GNN}&{ GCN/RNN }&{\checkmark}\\
    \rule{0pt}{9pt}
    {Know-Evolve(2022) \cite{knowEvolve}}&{Graph attention}&{RNN}&{--}&{Glean(2020)\cite{DBLP:conf/kdd/DengRN20}}&{GNN}&{GCN/GRU}&{\checkmark}\\
    \rule{0pt}{9pt}{RTFE(2022)\cite{rtfe}}&{Graph attention}&{Contrastive Learning}&{--}&{TeMP(2020)\cite{wu2020temp}}&{GNN}&{GCN/gating GRU}&{\checkmark}\\
    \rule{0pt}{9pt}{MPKD(2023)\cite{mpkd}}&{Graph attention}&{Knowledge Distillation}&{--}&{DACHA(2021)\cite{chen2021dacha}}&{GNN}&{GCN/Self-attention}&{\checkmark}\\
    \rule{0pt}{9pt}{CyGNet(2023)\cite{DBLP:conf/aaai/ZhuCFCZ21}}&{Graph attention}&{Knowledge Distillation}&{--}&{RE-GCN(2021)\cite{li2021temporal}}&{GNN}&{GCN/Self-attention }&{\checkmark}\\
    \rule{0pt}{9pt}{rGalT(2022)\cite{rgalt}}&{Transformer}&{self-attention}&{\checkmark}&{TANGO(2021)\cite{DBLP:conf/emnlp/HanDMGT21}}&{GNN}&{GCN/Neural ODE}&{\checkmark}\\
    \rule{0pt}{9pt}{GHT(2022)\cite{ght}}&{Transformer}&{self-attention}&{\checkmark}&{TiRGN(2022)\cite{DBLP:conf/ijcai/LiS022}}&{GNN}&{GCN/Double RNN }&{\checkmark}\\   
    \rule{0pt}{9pt}{HSAE(2023)\cite{hsae}}&{Transformer}&{self-attention}&{\checkmark}&{HiSMatch(2022)\cite{DBLP:conf/emnlp/LiHGJP0L0GC22}}&{GNN}&{GCN/RNN }&{\checkmark}\\
    \rule{0pt}{9pt}
    {KGFFP(2022)\cite{ge2022spatio}}&{Rule Mining}&{Manual Rule Mining}&{--}&{HGLS(2023)\cite{DBLP:conf/www/ZhangX0WW23}}&{GNN}&{GCN/Gating mechanism}&{\checkmark}\\
    \rule{0pt}{9pt}
    {TLogic(2022)\cite{liu2022tlogic}}&{Rule Mining}&{Temporal Random Walk}&{--}&{CluSTeR(2020)\cite{DBLP:conf/acl/LiJGLGWC20}}&{RL}&{ Beam-level rewards}&{--}\\
    \rule{0pt}{9pt}
    {TILP(2022)\cite{xiong2023tilp}}&{Rule Mining}&{Differentiable Learning}&{--}&{TAgent(2021)\cite{DBLP:conf/icassp/TaoLW21}}&{RL}&{ Binary terminal rewards }&{--}\\
    \rule{0pt}{9pt}
    {TFLEX(2022)\cite{lin2022tflex}}&{Rule Mining}&{Logical Embedding}&{--}&{TPath(2021)\cite{DBLP:journals/asc/BaiYCM21}}&{RL}&{Path diversity rewards }&{--}\\
    \rule{0pt}{9pt}
    {TPRG(2023)\cite{bai2023multi}}&{Rule Mining}&{Manual Rule Mining}&{--}&{TITer(2021)\cite{DBLP:conf/emnlp/SunZMH021}}&{RL}&{ Time-shaped rewards }&{--}\\
    \rule{0pt}{9pt}{TLmod(2023)\cite{bai2023temporal}}&{Rule Mining}&{Manual Rule Mining}&{--}&{DREAM(2023)\cite{DBLP:journals/corr/abs-2304-03984}}&{RL}&{ Attention/dynamic rewards}&{--}\\
    \rule{0pt}{9pt}
    {ALRE-IR(2023)\cite{mei2022adaptive}}&{Rule Mining}&{GRU}&{\checkmark}&{RLAT(2023)\cite{DBLP:journals/kbs/BaiC023}}&{RL}&{ LSTM/attention/RL }&{--}\\
\hline
\end{tabular}
\begin{tablenotes}
        \footnotesize              
        \item[1] \textbf{ED} represents whether TKGC methods are Encoder-Decoder structure, and draw \checkmark if it is; 
        \item[2] \textbf{Temporal Domain Gen} denotes \textit{Temporal Domain Generalization}.
      \end{tablenotes}
}
\end{threeparttable}
\label{tab:label_ext}
\end{table*}

\subsection{Reinforcement Learning-based TKGCs}
%Reinforcement learning (RL) is used to define and address the problem of learning a strategy to maximize reward or achieve a certain objective by a model during training. 
\textit{Reinforcement Learning (RL)} \cite{li2019reinforcement} adjusts the strategy based on feedback to maximize cumulative rewards in the process of interaction and finally obtain the optimal learning strategy. %\WangC{How about this sentence?} 
%\uline{RL-based TKGC methods generally introduce RL to ensure that the model has good training.}
RL-based methods treat TKGC as a \textit{Markov Decision Process (MDP)} \cite{DBLP:conf/ijcai/WanP00H20}, 
%\uline{The objective is to learn an optimal policy to infer missing elements in TKG. One of the key advantages of RL-based methods is their ability to} 
which is superior to producing explainable predictions. 

For instance, by establishing connections among the temporal events (\emph{COVID-19}, \emph{Infect}, \emph{Tom}, \emph{2022-12-3}), (\emph{Tom}, \emph{Talk\_to}, \emph{Fack}, \emph{2022-12-4}) and (\emph{Jack}, \emph{Visit}, \emph{City Hall}, \emph{2022-12-5}), 
RL-based TKGC methods can infer a new quadruplet (\emph{COVID-19}, \emph{Occur}, \emph{City Hall}, 
\emph{2022-12-6}) \cite{DBLP:journals/corr/abs-2304-03984}.
These methodes leverage historical TKG snapshots to infer answers for future-related queries. The components of the MDP are described as follows:

\begin{itemize}
\item \textbf{States.} Let $\mathcal{S}$ denote the state space, where a state is represented by a quintuple $s_l = \left(\mathbf{e}_l, \mathbf{t}_l, \mathbf{e}_q, \mathbf{t}_q, \mathbf{r}_q\right) \in \mathcal{S}$. Here, $\left(\mathbf{e}_l, \mathbf{t}_l\right)$ represents the node visited at step $l$, and $\left(\mathbf{e}_q, \mathbf{t}_q, \mathbf{r}_q\right)$ represents the elements in the query. The former represents local information, while the latter can be seen as global information. The agent starts from the source node of the query, so the initial state is $s_0 = \left(\mathbf{e}_q, \mathbf{t}_q, \mathbf{e}_q, \mathbf{t}_q, \mathbf{r}_q\right)$.

\item  \textbf{Observations.} An agent cannot observe the overall state of the environment. Intuitively, the answer remains hidden while the query and current location are visible to the agent. Formally, the observation function over the state is defined as $\mathcal{O}\left(\left(\mathbf{e}_l, \mathbf{t}_l, \mathbf{e}_q, \mathbf{t}_q, \mathbf{r}_q\right)\right)=\left(\mathbf{e}_l, \mathbf{t}_l, \mathbf{e}_q, \mathbf{r}_q\right)$.

\item \textbf{Actions.} Let $\mathcal{A}$ denote the action space, and $\mathcal{A}_l$ denote the set of available actions at step $l$. $\mathcal{A}_l \subset \mathcal{A}$ consists of the outgoing edges from node $\mathbf{e}_l^{t_l}$. More specifically, $\mathcal{A}_l$ should be $\left\{\left(\mathbf{r}', \mathbf{e}', \mathbf{t}'\right) \mid \left(\mathbf{e}_l, \mathbf{r}', \mathbf{e}', \mathbf{t}'\right) \in \mathcal{F}, \mathbf{t}' \leq \mathbf{t}_l, \mathbf{t}' < \mathbf{t}_q\right\}$, but since an entity often has multiple related historical events, this leads to a large number of possible actions. Therefore, the final set of available actions $\mathcal{A}_l$ are sampled from the aforementioned outgoing edges.
\item \textbf{Transition.} The environment state transitions to a new node through the edge selected by the agent. The transition function $\delta: \mathcal{S} \times \mathcal{A} \rightarrow \mathcal{S}$ is defined as $\delta\left(s_l, \mathcal{A}_l\right) = s_{l+1} = \left(\mathbf{e}_{l+1}, \mathbf{t}_{l+1}, \mathbf{e}_q, \mathbf{t}_q, \mathbf{r}_q\right)$, where $\mathcal{A}_l$ represents the sampled outgoing edges of $\mathbf{e}_l^{t_l}$.
\end{itemize}

%A significant challenge in applying RL to KG reasoning lies in constructing appropriate reward functions. 
When applying RL to the TKG reasoning task, a key challenge lies in constructing an appropriate reward function. Many existing methods rely on manually designed rewards.

\textbf{TAgent} \cite{DBLP:conf/icassp/TaoLW21} adopts binary terminal rewards for TKGC, which limits its ability to obtain sufficient rewards. In order to improve the quality of the reward function, subsequent methods have made novel attempts. \textbf{TPath} \cite{DBLP:journals/asc/BaiYCM21} introduces path diversity rewards, while \textbf{TITer} \cite{DBLP:conf/emnlp/SunZMH021} incorporates time-shaped rewards based on Dirichlet distribution to guide the model learning.
\textbf{CluSTeR} \cite{DBLP:conf/acl/LiJGLGWC20} utilizes the RNN to acquire temporal information and incorporates it into the beam-level reward function. However, these models heavily rely on manually designed rewards, which introduces limitations due to the sparse reward dilemma, laborious design process, and performance fluctuations. \textbf{DREAM} \cite{DBLP:journals/corr/abs-2304-03984} introduces an attention-based adaptive RL model to predict future missing items. The model consists of two main components: (1) a multi-faceted attention representation learning method that captures simultaneously semantic dependence and temporal evolution; (2) an adaptive RL framework that performs multi-hop reasoning by dynamically learning the reward function.

For temporal multi-hop reasoning,  
\textbf{TPath} \cite{DBLP:journals/asc/BaiYCM21} takes temporal information into consideration and selects specific multi-hop reasoning paths in TKGs. TPath proposes a policy network that can train the agent to learn temporal multi-hop reasoning paths. In addition, it also proposes a reward function that considers the diversity of temporal reasoning paths. 
\textbf{RLAT} \cite{DBLP:journals/kbs/BaiC023} combine RL with the attention mechanism for temporal multi-hop reasoning. RLAT uses LSTM and attention mechanism as memory components, which are helpful to train multi-hop reasoning paths. Second, an attention mechanism with an influence factor is proposed. This mechanism measures the influence of neighbor information and provides different feature vectors. The strategy function makes the agent focus on occurring relations with high frequency, allowing for multi-hop reasoning paths with higher correlation.

\section{Applications}\label{Sec:Applications}
In this section, we mainly summarize the applications of TKGC to some downstream tasks, including \textit{Question answering systems}, \textit{Medical and risk analysis systems} and \textit{Recommendation systems}.

\subsection{Question Answering Systems}
\begin{figure}[t]
    \begin{center}
    \includegraphics[width=0.95\columnwidth]{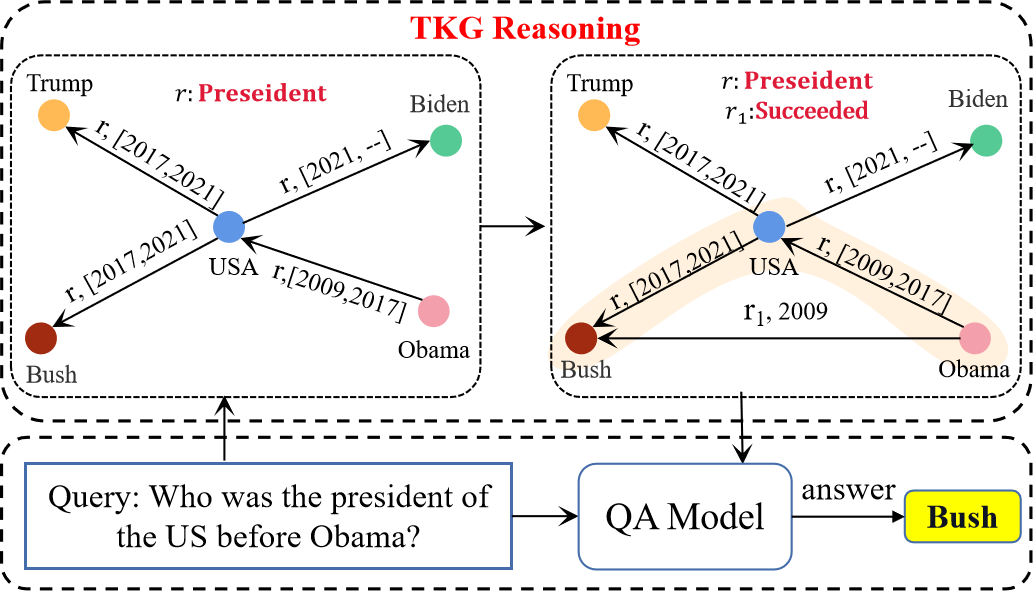}
    \end{center}
    \caption{The framework of question answering with TKG reasoning.}
    \label{fig:qa}
\end{figure}

Question answering systems are crucial applications of TKGCs. They typically perform queries and reasoning on TKGs, leveraging keywords from the question to predict missing entities or relations, and subsequently provide accurate answers to the questions. The specific process is shown in Fig. ~\ref{fig:qa}.

\textbf{Event-QA} \cite{eventqa} proposes a TKG designed for answering event-centric questions, including 1000 semantic queries and more than 970 thousand multilingual events. This method involves inferring missing entities or relations in the TKG using the keywords provided in the question, thereby guiding the QA model to provide an accurate answer.
In order to better address complex temporal reasoning questions based on the TKG. \textbf{TwiRGCN} \cite{twirgcn} designs a novel weighted GCN to answer questions that require complex temporal reasoning in TKGs.
Furthermore, \textbf{FORECASTTKGQA} \cite{forecasttkgqa} proposes a large-scale TKGQA dataset that aims to predict future facts. For each question in this dataset, the QA models can only have access to the TKG information before the timestamp annotated in the given question for answer inference.
\textbf{Ong}~\emph{et al.} \cite{ong} propose a new TKGQA dataset based on the TKGC task,which associates over 5000 financial news documents with question-answer pairs.

Previous studies have identified various issues, notably the omission of specific time references within the TKG and the neglect of the temporal order of timestamps. To address these issues, \textbf{TSQA} \cite{tsqa} uses the timestamp estimation module to infer the timestamp of the question, and uses a time-sensitive KG encoder to fuse ordering information into TKG embedding.
The existing question-answer scheme on the TKG mainly focuses on a simple temporal question, which can rely on a single TKG fact. \textbf{TempoQR} \cite{tempoqr} proposes a framework to solve complex questions by retrieving relevant information from the underlying TKG based on the keywords in question. Afterwards, this method infers temporal information through the TKGC task, eliminating the necessity of directly accessing the TKG. Likewise, \textbf{CTRN} \cite{ctrn} captures implicit temporal and relation representations of each question via the TKG reasoning process, and then generates accurate answers through the QA model.

\subsection{Medical and Risk Analysis Systems}
Medical and risk analysis systems are important applications of TKGC, e.g., medical diagnosis systems \cite{song2020research} and risk analysis systems \cite{chen2021spatio}. 

For the medical domain, traditional systems are designed based on static data, which is difficult to reflect the dynamical variation characteristics of data. \textbf{Song}~\emph{et al.} \cite{song2020research} first explore GRU to integrate the temporal information into the KG, and then apply TransR \cite{Lin2015learning} to ensure the structural completeness of the TKG. Finally, they improve the accuracy of medical diagnosis systems by leveraging the complete TKG. \textbf{Yang} ~\emph{et al.} \cite{yang2020temporal} propose a novel Chinese medical search system that applies the TKG to represent the dynamic changing of traditional Chinese medicine. Afterwards, they propose a TKGC model to complete the temporal intentions of search sentences for medical diagnosis.

Since most data, in reality, is multi-source spatio-temporal data, it is difficult for traditional static data to reflect the temporal dependency among the data. Especially for analysis systems, most of the data involved are time-dependent, such as weather data and traffic data. \textbf{Lin} ~\emph{et al.} \cite{zhao2020urban} first present a TKGC model and utilize the complete TKG for analyzing the multi-source data in smart cities. Likewise, \textbf{KG4MR} \cite{chen2021spatio} designs a TKGC model to predict relations between risky weather and models the relationship among risky weather events, human activity events and element attributes in knowledge graphs. Afterwards, KG4MR proposes a query method to solve the spatio-temporal intersection reasoning and successfully applied it to the Olympic Winter Games Beijing $2022$.

\subsection{Recommendation Systems}
\begin{figure}[t]
    \begin{center}
    \includegraphics[width=0.95\columnwidth]{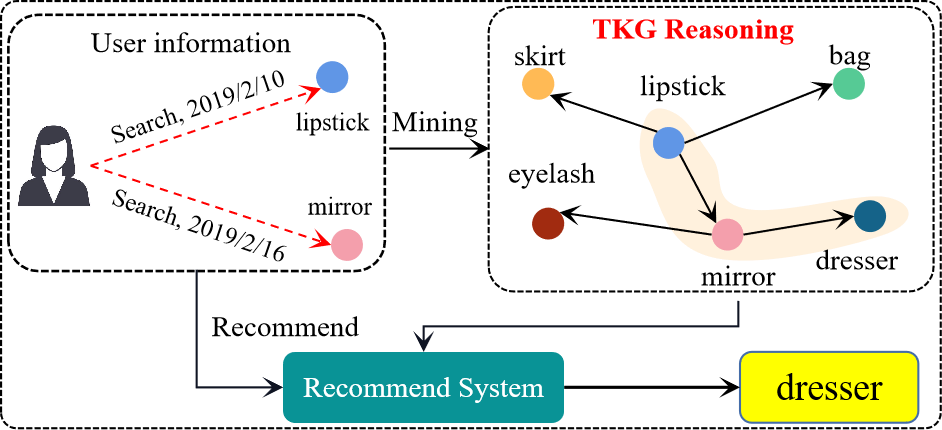}
    \end{center}
    \caption{The framework of recommendation with TKG Reasoning.}
    \label{fig:re}
\end{figure}

Recommendation systems based on TKGC mainly analyze the historical behaviors and preferences of users, and subsequently leverage entity-related TKGs to identify the relevant products or services that users may purchase. The framework is shown in Fig. ~\ref{fig:re}.

\textbf{TSTKG4Rec} \cite{tstkg} constructs a rich travel spatial-temporal KG derived from diverse sources including Baidupedia, Interactive Encyclopedia and Wikipedia. It addresses the problem of fusion of multi-source heterogeneous data and lacking of spatial-temporal information.
In order to update the representation of related entities, \textbf{Xiao}~\emph{et al.} \cite{xiao} propose an incremental construction model for TKGs, which emphasizes the importance of semantic path information between involved entities and the interaction in refining their representations. Specifically, this model extracts semantic paths of varying lengths between user and item, and captures the semantic information of the path and interaction itself information through RNN to update entity representations. 
In addition, most recommendation systems provide unchanged services without making corresponding improvements over time. To address this issue, \textbf{TASR} \cite{tasr} provides service recommendations by leveraging the knowledge graph of practice. It incorporates multiple and complex interactions between heterogeneous entities, including modeling the user-service interactions over time through the construction of a \textit{Temporal Service Knowledge Graph (TSKG)}. TASR explores the TSKG and extracts top-rated services to enhance the quality of service recommendations.
With regard to the challenges of representing users' dynamic mobility behaviors and modeling users' long- and short-term preferences using TKGs, \textbf{STKGRec} \cite{stkgrec} constructs a spatial-temporal KG based on users' historical check-in sequence. It enables the promotion of the next \textit{Point-of-Interest (POI)} recommendations without introducing external attribute information of users and POIs.

\subsection{Others}

TKGC also has been widely applied in many other applications, e.g., citiation prediction \cite{zong2022citation}, and mobility prediction \cite{wang2021spatio}. \textbf{STKG} \cite{wang2021spatio} models the urban mobility trajectories as a temporal knowledge graph, where mobility trajectories, category information of venues, and temporal information are jointly modeled by the facts with different relation types. Then, the mobility prediction is converted to the TKGC problem through an embedding model that captures spatio-temporal patterns. \textbf{CTPIR} \cite{zong2022citation} proposes a citation trajectory prediction framework that captures the dynamic influence of citation to predict the future citation trajectory of a paper. It first adopts the R-GCN model to capture the connections between two snapshots. Then, it learns a fine-grained influence representation for trajectory prediction.

\section{Future Directions}\label{Sec:Future}
In the previous sections, we comprehensively reviewed the TKGC literature and provided an in-depth dissection of them. Meanwhile, we also identified many challenges and open problems that need to be addressed.
In this section, we discuss the future directions of this research area.

%\subsection{Flexibility of the Temporal Regularization}
%Temporal regularization is an important component to ensure facts behave smoothly over time in TKG. Most of the existing temporal regularization methods generally commonly execute a subtraction operation on neighboring timestamps to achieve temporal smoothness. Specifically, they focus more on the element level, requiring matching elements of corresponding temporal embeddings to be close or even identical. In addition, only one method focuses on the structural level. In particular, it explores cosine similarity to measure the global similarity between neighboring timestamps. Both strategies can increase time smoothness to some extent, but their flexibility is severely limited. In the future direction, the structure level-based temporal regularization should be given more attention and the introduction of bias to further enhance flexibility. In addition, more advanced temporal regularization should be further investigated.

\subsection{Multi-Modal Temporal Knowledge Graphs}
The rapid development of Internet technology has given rise to the emergence of various forms of data. However, a single modality is often insufficient to fully represent an object, and leveraging multiple modalities for joint representation can enrich the semantics from diverse perspectives. Similarly, this holds true for KGs. Most of the existing KGs consist solely of textual information, which often fails to provide a comprehensive and detailed description of entities, resulting in limited expression of accurate and rich semantics. Consequently, the development of multi-modal KGs becomes increasingly crucial as they allow for diverse representations of entities, enabling a more nuanced and enriched understanding of their semantics. Of course, several research teams have initiated the exploration and construction of multi-modal static KGs. However,  the static KGs cannot reflect the temporal correlation and the evolution process. In contrast, multi-modal TKGs can describe the rich semantics of entities, the temporal evolution of events, and the semantic, temporal, and spatial relationships between entities. This enriched representation can be leveraged more effectively in downstream tasks, such as visual question answering and recommendation systems.

\subsection{Inductive and Few-Shot Learning Settings} 

The majority of TKGC methods typically rely on the assumption that a large number of training examples are available to learn entity representations, i.e., transductive settings. These methods typically aim to complete the missing facts in TKGs by leveraging the known information in the graph. However, in real-world settings, TKGs often exhibit long-tail distributions, meaning that there are many rare entities and relations (i.e., few-shot settings) or even unseen entities and relations (i.e., inductive settings). These unseen entities and relations pose a challenge for traditional TKGC methods, as they lack sufficient training data to learn accurate representations. In such scenarios, traditional methods may struggle to provide optimal representations for these rare entities and relations. The sparsity of data makes it difficult to generalize and make accurate predictions for these entities. As a result, these methods may fail to effectively complete the missing facts associated with unseen or rare entities.

\subsection{ Logical Query Answering via Temporal Conditions} 

Answering complex queries with temporal information, e.g., \emph{what was the first film Julie Andrews starred in after her divorce with Tony Walton}, is an interesting direction for question answerings. Complex query answering has been extensively studied in static KGs, static KGC methods map queries into the vector space and model logical connectives (conjunction, disjunction, and negation) as neural geometric operations. However, these methods do not apply to queries with temporal conditions.
Future work can focus on developing more sophisticated complex query embedding models that can handle complex temporal conditions, such as temporal intervals, durations, and granularities. This can enable more accurate and precise query answering, especially when dealing with temporal constraints and evolving information in TKGs.

\subsection{Unification with Large Language Models}

\textit{Large Language Models (LLMs)} have shown great performance in various applications. Pre-trained on the large-scale corpus, LLMs enable to contain enormous general knowledge and reasoning ability. Recently, the possibility of unifying LLMs with KGs has attracted increasing attention from researchers and practitioners \cite{pan2023unifying}. Much research has utilized LLMs to tackle tasks in the field of KGs. However, the unification of LLMs with TKGC methods is less explored by existing research. LLMs are pre-trained on the static corpus which are inadequate in capturing the temporal information. Moreover, TKGs are evolving over time with new knowledge added. How to enable LLMs effectively model the dynamic dependence of TKG and represent new knowledge is still an open question.

\subsection{Interpretability Analysis} 
Knowledge graphs are credited for their good interpretability. However, most existing TKGC methods are based on deep learning algorithms which are black-box models. The reasoning process of TKGC methods used to arrive at their results is not explainable to humans. This largely limits their applications in high-stake scenarios, such as medical diagnosis and legal judgment. Although some works \cite{han2021explainable, jung2021learning, du2023imf} attempt to provide human-understandable evidence explaining the forecast,
%For example, rule-based TKGC methods provide explanations in the form of temporal logical rules \cite{liu2022tlogic,xiong2023tilp,}.
they focus on simple models with plain explanations e.g., logical rules and paths. Explaining the complex captured temporal patterns utilized for reasoning and interpreting the more intricate TKGC models still remains an unresolved matter.

% {\color{red}{Thus, the interpretability of TKGC methods aims to understand the reasoning process and provide a human-understandable explanation. Although, the rule-based methods can  }}

\section{Conclusion}\label{Sec:Conclusion}
\textit{Temporal Knowledge Graph Completion (TKGC)} is an emerging and active research direction that has attracted increasing attention from both academia and industry. In this paper, we presented a comprehensive overview of the recent research in this field. Firstly, we detailed the interpolation methods and further categorized them based on how they handle temporal information. The extrapolation methods were then further described and classified based on how they predict future events. Finally, we discussed the challenges and future directions in this field. %We believe that this paper will provide a thorough overview of this topic and help to advance future research.

\balance

\bibliographystyle{IEEEtran}
\bibliography{Survey}
\vfill

\end{document}